\documentclass{article}

\usepackage{microtype}
\usepackage{graphicx}
\usepackage{subfigure}
\usepackage{booktabs} % for professional tables

\usepackage{hyperref}

\usepackage[accepted]{icml2025}

\usepackage{amsmath}
\usepackage{amssymb}
\usepackage{mathtools}
\usepackage{amsthm}
\usepackage{enumitem}
\usepackage{bbm}
\usepackage{multirow}
\usepackage{array}
\DeclareMathOperator*{\argmin}{arg\,min}

\usepackage[capitalize,noabbrev]{cleveref}

\theoremstyle{plain}
\newtheorem{theorem}{Theorem}[section]

\theoremstyle{definition}
\newtheorem{definition}[theorem]{Definition}

\theoremstyle{remark}

\usepackage[textsize=tiny]{todonotes}

\icmltitlerunning{A Neuro-inspired Interpretation of Unlearning in Large Language Models through Sample-level Unlearning Difficulty}

\begin{document}

\twocolumn[
\icmltitle{A Neuro-inspired Interpretation of Unlearning in Large Language Models through Sample-level Unlearning Difficulty}




\begin{icmlauthorlist}
\icmlauthor{Xiaohua Feng}{xxx}
%\icmlauthor{Firstname2 Lastname2}{equal,yyy,comp}
\icmlauthor{Yuyuan Li}{yyy}
\icmlauthor{Chengye Wang}{xxx}
\icmlauthor{Junlin Liu}{xxx}
\icmlauthor{Li Zhang}{xxx}
\icmlauthor{Chaochao Chen}{xxx}
%\icmlauthor{Firstname8 Lastname8}{sch}
%\icmlauthor{Firstname8 Lastname8}{yyy,comp}
\end{icmlauthorlist}

\icmlaffiliation{xxx}{Zhejiang University, Hangzhou, China}
\icmlaffiliation{yyy}{Hangzhou Dianzi University, Hangzhou, China}
%\icmlaffiliation{zzz}{School of ZZZ, Institute of WWW, Location, Country}

\icmlcorrespondingauthor{Chaochao Chen}{zjuccc@zju.edu.cn}
%\icmlcorrespondingauthor{Firstname2 Lastname2}{first2.last2@www.uk}

\icmlkeywords{Machine Learning, ICML}

\vskip 0.3in
]



\printAffiliationsAndNotice{\icmlEqualContribution} % otherwise use the standard text.

\begin{abstract}

%Large-scale language models (LLMs) excel in various generative tasks but face critical challenges, such as privacy leakage and biases. To address these issues under legal and regulatory frameworks, LLM unlearning has emerged as a potential solution. 
Driven by privacy protection laws and regulations, unlearning in Large Language Models (LLMs) is gaining increasing attention.
However, current research often neglects the interpretability of the unlearning process, particularly concerning sample-level unlearning difficulty.
Existing studies typically assume a uniform unlearning difficulty across samples.
This simplification risks attributing the performance of unlearning algorithms to sample selection rather than the algorithm's design, potentially steering the development of LLM unlearning in the wrong direction.
Thus, we investigate the relationship between LLM unlearning and sample characteristics, with a focus on unlearning difficulty. 
Drawing inspiration from neuroscience, we propose a Memory Removal Difficulty ($\mathrm{MRD}$) metric to quantify sample-level unlearning difficulty. 
Using $\mathrm{MRD}$, we analyze the characteristics of hard-to-unlearn versus easy-to-unlearn samples. %, e.g., sample occurrence frequency, semantic complexity, and other factors.
%\lyy{add findings, and therefore we enhance ...}
%
%These characteristics significantly influence the MRD values of samples, therefore we enhance the widely used Stochastic Gradient Ascent (SGA) method by introducing Curriculum Gradient Ascent (CGA), 
Furthermore, we propose an $\mathrm{MRD}$-based weighted sampling method to optimize existing unlearning algorithms, which prioritizes easily forgettable samples, thereby improving unlearning efficiency and effectiveness.
We validate the proposed metric and method using public benchmarks and datasets, with results confirming its effectiveness.
%
%This findings offer new insights into LLM unlearning and pave the way for further algorithmic improvements.\lyy{can be simplified}
\end{abstract}

\section{Introduction}

Large Language Models (LLMs) excel at generating human-like text, leading to their broad adoption in various applications.
This success largely stems from their strong memorization of the training corpus~\cite{zhang2023counterfactual}.
However, such memorization also raises serious concerns, including risks of privacy breaches~\cite{kim2024propile}, bias propagation~\cite{yu2023unlearning,motoki2024more}, and the generation of illegal content~\cite{karamolegkou2023copyright}.
In particular, privacy protection laws like the \textit{GDPR} require service providers to remove private information from training data upon user request~\cite{voigt2017eu}.
This creates a significant challenge: how to effectively erase the influence of specific data samples (i.e., \textit{the forget set}), or higher-level data concepts from pre-trained LLMs.

A practical approach to addressing the issue above is Machine Unlearning (MU)~\cite{liu2024rethinking}. 
Previous research~\cite{ginart2019making,ullah2021machine,thudi2022unrolling,liu2024model} has primarily focused on MU in classification models, where retraining on the remaining data (i.e., \textit{the retain set}) is the gold standard. 
However, given the massive scale of training data and the extensive number of parameters in LLMs, this unlearning approach becomes infeasible for LLMs. 
Therefore, developing effective and efficient methods for implementing MU in LLMs represents a critical challenge that requires resolution.

Existing studies~\cite{jang2023knowledge,ji2024reversing,feng2024fine,liu2024rethinking} defines LLM unlearning as the removal of specific knowledge from the forget set (i.e., \textit{unlearning completeness}) while preserving the model's performance on unrelated tasks (i.e., \textit{model utility}).
Current methods achieving this can be broadly classified into three categories, i.e., gradient-based methods~\cite{jang2023knowledge,yao2024large}, preference optimization-based methods~\cite{maini2024tofu,zhang2024negative}, and model weight-based methods~\cite{jia2024wagle}.
%
% \begin{itemize}[leftmargin=*] \setlength{\itemsep}{-\itemsep}
%     \item \textbf{Gradient-based methods}. These include gradient ascent techniques directly applied to the forget set, as well as variants with added regularization terms.
%     \item \textbf{Preference optimization-based methods}. These treat the forget set as negative examples in preference alignment or assign predefined responses (e.g., refusal content) to the forget set to achieve unlearning.
%     \item \textbf{Model weight-based methods}. These focus on leveraging the modular structure of LLMs to identify the roles of different components, providing targeted guidance for unlearning.
% \end{itemize}
%
%Despite the significant progress made in unlearning algorithms for LLMs in recent years, research on the interpretability of the unlearning process in LLMs remains relatively underdeveloped. 
%
Despite recent advancements, the interpretability of the unlearning process in LLMs remains underexplored.
\textit{The lack of interpretability hinders the capability to comprehensively evaluate the practical effectiveness of existing LLM unlearning algorithms.}
For instance, the superior performance of certain unlearning algorithms might be attributed merely to the inherent ease of unlearning the selected samples, rather than to any genuine advantage of the algorithms themselves. 
Such a lack of fine-grained analysis could potentially impact the reliability and generalizability of LLM unlearning algorithms.

Recent studies increasingly explore the interpretability of MU.
For example, \citet{fan2024challenging} analyze how different partitions of the forget sets influence model performance on the retain sets in image classification tasks. 
\citet{zhao2024makes} investigate the presence of explainable features within the forget sets and their impact on the difficulty of unlearning. 
\citet{chen2024cure4rec} provide a more fine-grained perspective, showing that in recommendation systems, unlearning difficulty varies significantly across users, with potential implications for the evaluation of unlearning algorithms. 
Collectively, these studies highlight a trend toward sample-level analysis in unlearning interpretability.
However, notable limitations remain. 
These works lack a formal definition of unlearning difficulty at the sample level and offer little theoretical insight into why certain samples are harder to unlearn. 
Additionally, methods developed for image classification may not effectively generalize to LLMs, which struggle with modeling structured features due to their text-based autoregressive nature. 
To address these issues, this paper investigates the LLM unlearning problem, focusing on the following three key questions:
\begin{itemize}[leftmargin=*] \setlength{\itemsep}{-\itemsep}
    %[leftmargin=*]
    %\setlength{\itemsep}{0pt}
    \item \textbf{Q1}. How to design a reasonable and computationally efficient metric to measure the unlearning difficulty of individual data samples?
    \item \textbf{Q2}. Based on the proposed metric, what characteristics make certain data samples more difficult to unlearn?
    \item  \textbf{Q3}. Can this metric enhance the effectiveness and efficiency of existing LLM unlearning algorithms?
\end{itemize}

\begin{figure}[t]
\vskip 0.2in
\begin{center}
\centerline{\includegraphics[width=\columnwidth]{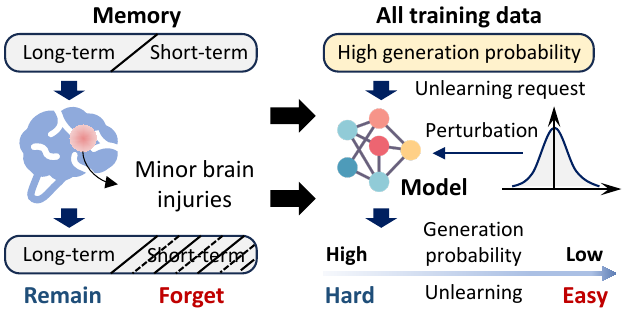}}
\caption{%Definition of unlearning difficulty. 
Unlearning difficulty is measured by introducing small perturbations to model parameters (akin to minor brain injuries) and comparing the change in generation probability for a specific sample before and after perturbation. A small change indicates the sample resides in the model's long-term memory and is harder to unlearn, whereas a large change suggests easier unlearning.}
\label{fig:illustrate}
\end{center}
\vskip -0.2in
\end{figure}

%Addressing these questions will advance our understanding of unlearning mechanisms in LLMs, establish a robust theoretical basis for developing more precise and efficient unlearning algorithms, and offer practical guidance for their implementation.
%
To address the questions above, this paper undertakes the following contributions:

\textbf{To address Q1,} we propose a metric, Memory Removal Difficulty ($\mathrm{MRD}$), to measure the unlearning difficulty of individual samples (e.g., sentences) in LLMs.
Inspired by findings in neuroscience~\cite{kim1992modality,squire1995retrograde,frankland2005organization,konrad2011long}, where long-term memories in the human brain are typically resistant to minor brain injuries and are not easily forgotten, $\mathrm{MRD}$ models unlearning difficulty in LLMs.
As shown in Figure~\ref{fig:illustrate}, it is formally defined as the expected change in the log-likelihood of a data sample before and after random perturbations to model parameters, ensuring both reasonable and computational feasibility.

\textbf{To address Q2,} we conduct an in-depth discussion on the $\mathrm{MRD}$ metric to uncover the characteristics of data samples that make them more difficult to unlearn. 
For instance, we find that samples with high frequency or those with strong contextual associations to other samples are often harder to unlearn. 
Through theoretical analysis and experimental validations, we provide clear explanations for these properties, thereby offering insights into the factors influencing the unlearning difficulty of individual data samples.

\textbf{To address Q3,} we propose an $\mathrm{MRD}$-based weighted sampling method to optimize existing unlearning algorithms. 
Inspired by curriculum learning, $\mathrm{MRD}$ serves as a scoring function to adjust the sampling probability of unlearning samples, enabling a dynamic progression from simple to complex unlearning sequences. 
%
%Comparative experiments show that the proposed method markedly accelerates convergence and enhances performance.
%
%This demonstrates that $\mathrm{MRD}$ effectively captures unlearning difficulty and provides a practical framework for improving unlearning algorithms.
%
Comparative experiments demonstrate that this method significantly accelerates convergence and improves performance, highlighting $\mathrm{MRD}$ as an effective measure of unlearning difficulty and a practical tool for optimizing unlearning algorithms.

%Looking ahead, we plan to investigate additional $\mathrm{MRD}$-based methods to further improve the accuracy and efficiency of unlearning algorithms, addressing practical demands in LLM applications. 
%
%This study offers key insights into LLM unlearning mechanisms and establishes a strong theoretical foundation for developing more effective and efficient unlearning strategies.

% \begin{itemize}[leftmargin=*] \setlength{\itemsep}{-\itemsep}
% %\begin{itemize}[leftmargin=*]
%     %\setlength{\itemsep}{0pt}
%     \item We propose a new metric $\mathrm{MRD}$ to quantify the unlearning difficulty of data points in LLMs, which is theoretically grounded, computationally efficient, and designed specifically from the perspective of LLM unlearning.
%     \item Using the $\mathrm{MRD}$ metric, we analyze which data points are more difficult to unlearn in LLMs and explain this phenomenon by linking $\mathrm{MRD}$ to key data characteristics.
%     \item We leverage $\mathrm{MRD}$ to enhance existing unlearning algorithms. By employing $\mathrm{MRD}$ as a scoring function in curriculum learning, we replace random unlearning with curriculum unlearning, which achieves faster convergence in theory.
%     \item We conduct extensive experiments on multiple real-world datasets and mainstream LLMs. Results confirm that $\mathrm{MRD}$ effectively captures unlearning difficulty, providing a solid basis for optimizing unlearning algorithms.
% \end{itemize}

\section{Related Work}

\subsection{Machine Unlearning}

MU methods can be categorized into exact unlearning and approximate unlearning~\cite{xu2023machine}.
Exact unlearning methods treat the retrained model as the gold standard to achieve complete erasure of the target data, i.e., aiming for 100\% unlearning completeness. 
These methods divide the model or dataset into multiple sub-components and construct an ensemble system, thereby distributing the computational overhead of retraining across these sub-components during the unlearning process~\cite{bourtoule2021machine, li2024ultrare}.
In contrast, approximate unlearning methods aim to obtain a model that is approximately equivalent to the retrained model in terms of either model parameters or outputs. 
These methods are typically achieved by estimating the influence of the target data~\cite{koh2017understanding, liu2024certified} or by fine-tuning a defined objective function.

\subsection{LLM Unlearning}

LLM unlearning is typically framed as approximate unlearning, aiming to achieve both high unlearning completeness and model utility. 
\citet{jang2023knowledge} first propose a gradient ascent method on the forget set, which significantly improves unlearning completeness but at the cost of reduced model utility.
To mitigate this, subsequent studies~\cite{maini2024tofu,yao2024large} introduce regularization-based enhancements (e.g., parameter and loss regularization). 
However, these methods still face challenges in balancing the trade-off between completeness and utility.
Later studies~\cite{zhang2024negative} approach unlearning by treating the forgotten data as negative examples in preference alignment, formalizing the process as a preference optimization task with predefined positive responses (e.g., refusals or counterfactual samples).
While this integrated optimization approach has shown some success, it suffers from low unlearning efficiency, limiting its practicality.
Recent research~\cite{jia2024wagle} revisits the problem through model weights, leveraging the modular structure of LLMs to identify and guide unlearning at the module level.
Although this method provides valuable insights, its computational efficiency remains low, posing significant challenges for real-world applications.
\section{Interpretability of LLM Unlearning}

This section establishes the LLM unlearning problem and defines unlearning difficulty.
We further analyze sample characteristics influencing unlearning difficulty and propose effective algorithmic improvements.

\subsection{Problem Setup of LLM Unlearning}

\paragraph{Autoregressive Model Training.} Given a training set $\mathcal{D} = \mathcal{D}_{F} \cup \mathcal{D}_{R}$, where
$\mathcal{D}_F = \{\boldsymbol{x}^1, \boldsymbol{x}^2, \ldots, \boldsymbol{x}^{N_f}\}$ and $\mathcal{D}_R = \{\boldsymbol{x}^1, \boldsymbol{x}^2, \ldots, \boldsymbol{x}^{N_r}\}$ represent the forget and retain sets with $N_f$ and $N_r$ samples, respectively, each sample $\boldsymbol{x}^i = \{x_1, \ldots, x_{n_i}\}$ corresponds to a sample of length $n_i$.
The parameters $\boldsymbol{\theta}'$ of a model autoregressively trained on $\mathcal{D}$ satisfy the following equation: 
\begin{align}
    \boldsymbol{\theta}' 
    &= \argmin_{\boldsymbol{\theta}} \mathcal{L}_{NLL} (\mathcal{D}; \boldsymbol{\theta}) \notag \\
    &= \argmin_{\boldsymbol{\theta}} -\mathbb{E}_{\boldsymbol{x}^i \sim \mathcal{D}} 
    \left[ \sum_{t=1}^{n_i} \log p(x_t \mid \boldsymbol{x}_{<t}; \boldsymbol{\theta}) \right].
    \label{eq:auto-reg}
\end{align}

\paragraph{Objective of LLM Unlearning.} To unlearn a sample $\boldsymbol{x}^i$, the objective is typically formalized as the following optimization problem~\cite{jang2023knowledge,ji2024reversing,jia2024wagle,liu2024rethinking}:
\begin{align}
    \max_{\boldsymbol{\theta}} \quad & \frac{1}{N_r} \sum_{g \in G} \sum_{\boldsymbol{x}_r \in \mathcal{D}_R}  g(\boldsymbol{x}_r; \boldsymbol{\theta}) \notag \\
    \text{subject to} \quad & \frac{1}{N_f} \sum_{\boldsymbol{x}_f \in \mathcal{D}_F}  \psi(\boldsymbol{x}_f; \boldsymbol{\theta}) \geq \epsilon,
\end{align}
where $\psi(\mathcal{D}_F; \boldsymbol{\theta})$ quantifies unlearning completeness, $G$ is a set of functions assessing other model capabilities (i.e., model utility), and $\epsilon$ is a threshold.
For example, $\psi(\mathcal{D}_F; \boldsymbol{\theta})$ can evaluate whether the model's memory of $\mathcal{D}_F$ is erased (e.g., by ensuring the probability of generating $\mathcal{D}_F$ is below $\epsilon$, or the divergence between the model's output distribution on $\mathcal{D}_F$ and the true distribution exceeds $\epsilon$). 
Meanwhile, $g(\mathcal{D}_R; \boldsymbol{\theta})$ assesses retained capabilities, such as minimizing the divergence between the model's output distribution on $\mathcal{D}_R$ and the true distribution.
In summary, the objective is to satisfy the unlearning constraints while minimizing degradation to the model’s other capabilities.

\subsection{Motivation}

\paragraph{Impact of Sample Selection on Unlearning Evaluation.} Most studies~\cite{maini2024tofu,li2024wmdp,liu2024rethinking} evaluate unlearning algorithms using random data unlearning, where the forget set is randomly drawn from the training set.
Performance is assessed based on unlearning completeness and the utility of the updated model.
However, random sample selection can lead to substantial performance variability across LLM unlearning methods, compromising the fairness of comparisons.

\begin{figure*}[htbp]
    \centering
    \subfigure[Retain Set]{
        \includegraphics[width=0.31\linewidth]{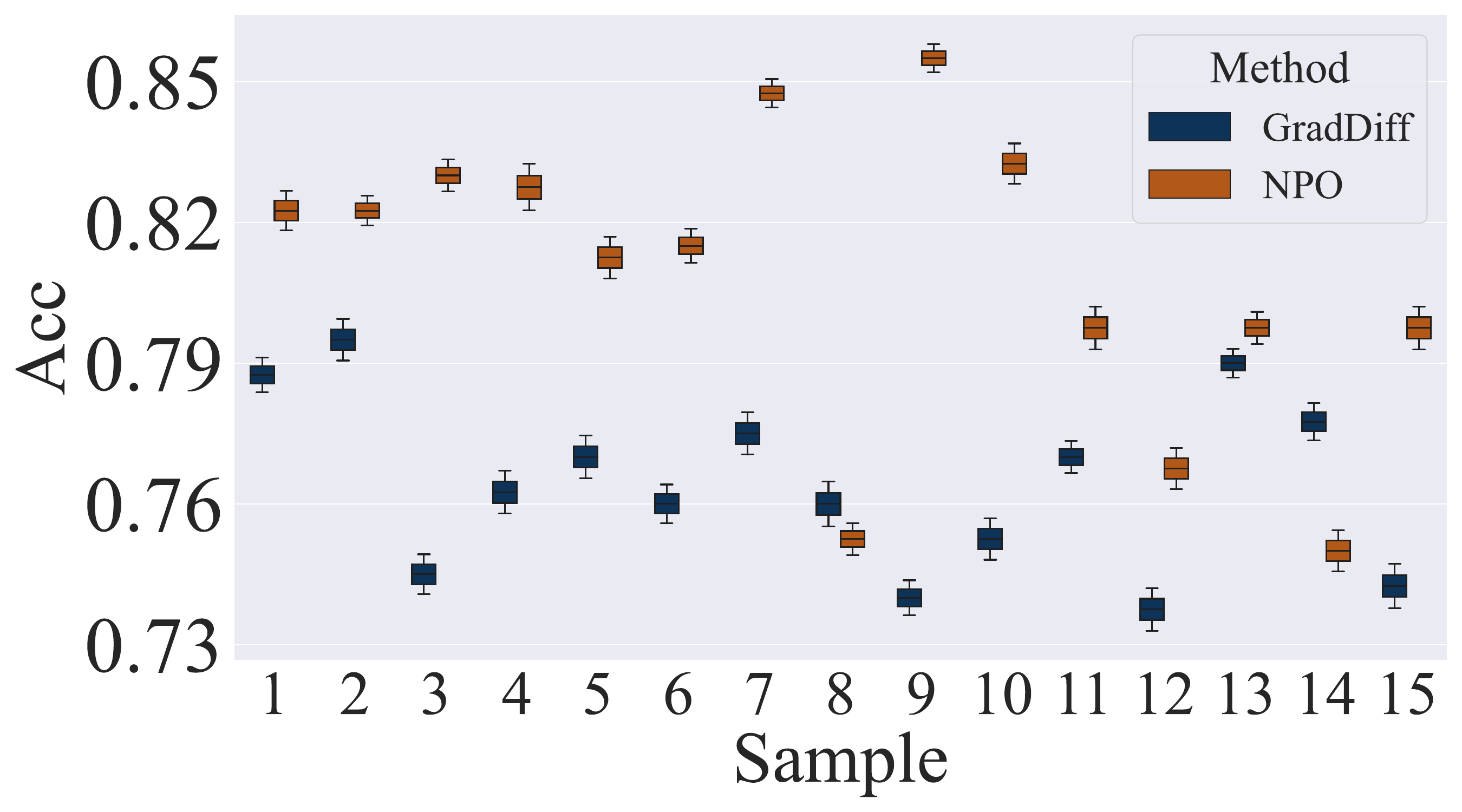} 
    }
    \subfigure[Real Author]{
        \includegraphics[width=0.31\linewidth]{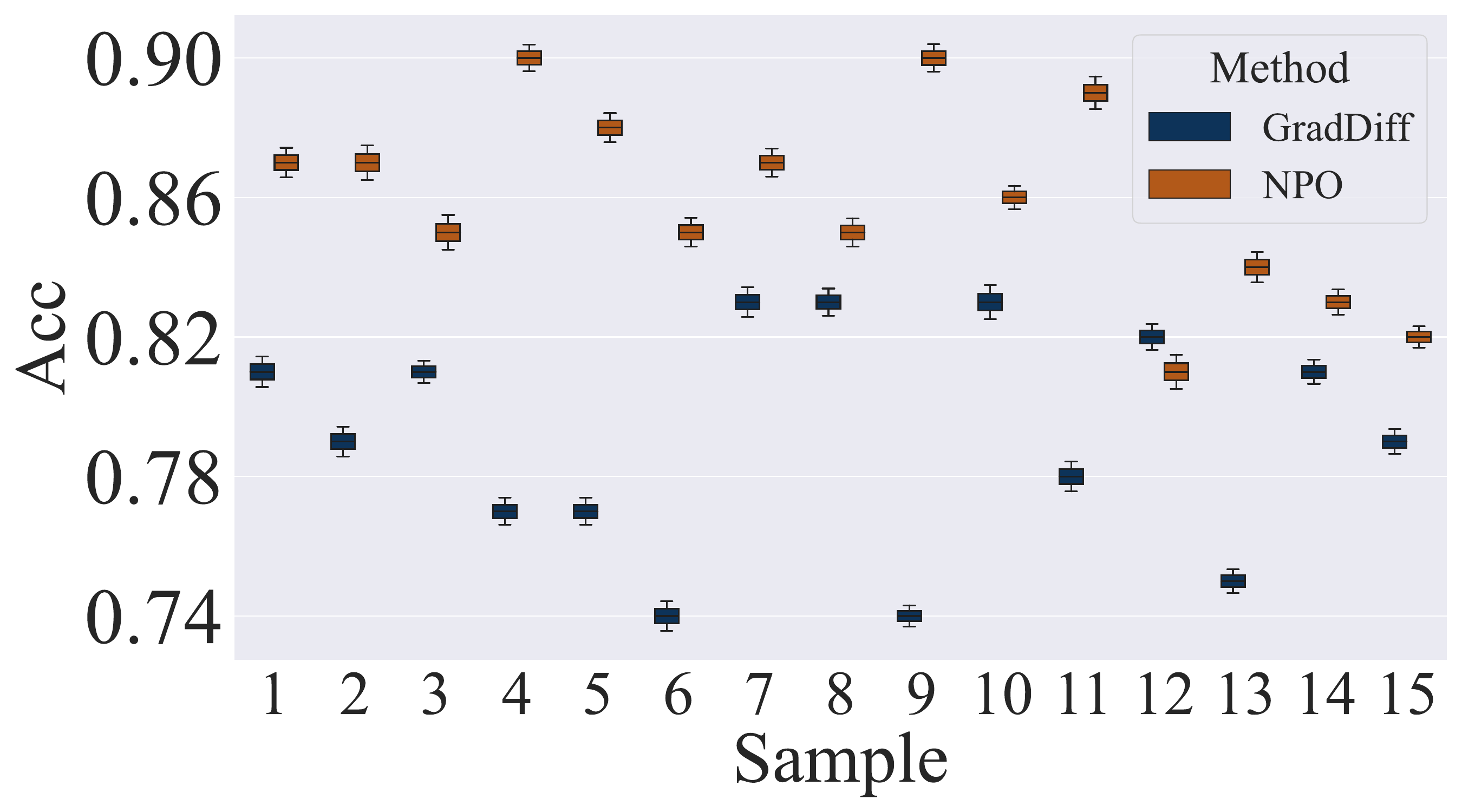} 
    }
    \subfigure[World Fact]{
        \includegraphics[width=0.31\linewidth]{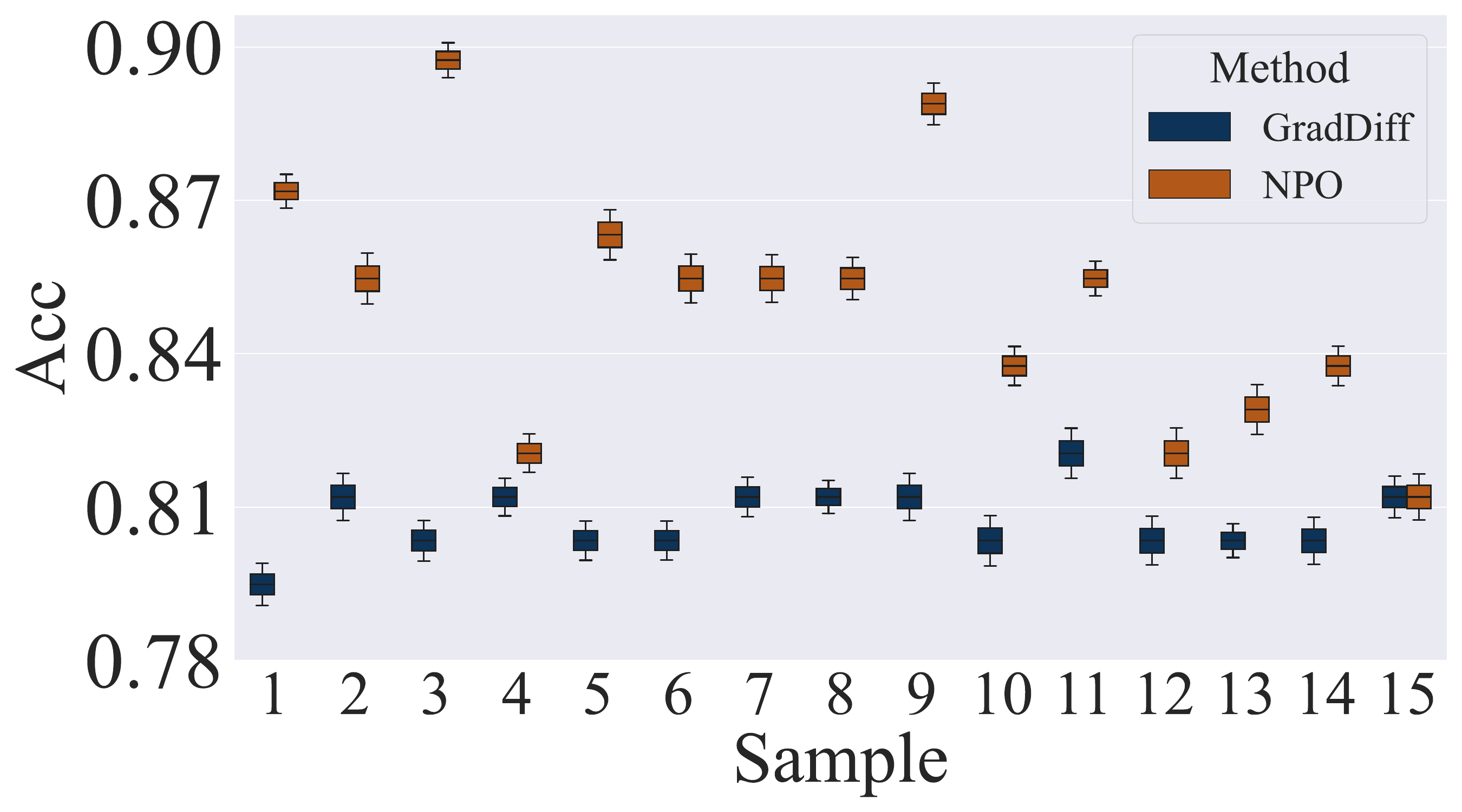} 
    }
    \caption{Impact of sample selection on unlearning evaluation. We report the variability in performance across different LLM unlearning methods (GradDiff and NPO).}
    \label{fig:motivation}
\end{figure*}

To investigate this, we analyze two mainstream LLM unlearning methods through systematic experiments on widely used benchmark datasets. 
Following prior studies, we impose uniform unlearning constraints, requiring the MA (Appendix~\ref{app:con-ear}) on unlearned samples to fall below a specified threshold as the termination condition. 
To account for existing methods, we evaluate both single-sample and group-sample unlearning scenarios. 
In each case, unlearning samples are randomly selected, and experiments are repeated five times to compare performance. 
Results presented in Figure~\ref{fig:motivation} highlight the uncertainties caused by random sample selection and its impact on method comparisons.%, where the direction of the points is random.

Specifically, we reveal two key observations from Figure~\ref{fig:motivation}.
First, for the same unlearning algorithm, the mean performance of the model varies significantly after unlearning different samples, indicating that selecting different unlearning samples leads to significant variance in unlearning effectiveness. %, with this variance increasing as the number of unlearning samples grows\lyy{how can this be observed?}. 
Second, it can be observed that the model's performance in NPO significantly outperforms GradDiff when unlearning most samples. 
However, there are certain samples for which GradDiff outperforms NPO after unlearning, indicating that the ranking of unlearning effectiveness among algorithms may reverse depending on the choice of unlearning samples.
%
%Although averaging performance across five random trials can approximate unlearning effectiveness, considerable bias remains.\lyy{?}

% \begin{figure}[htbp]
%     \centering
%     \subfigure[Retain Set]{
%         \includegraphics[width=\columnwidth]{figure/lab2_1.png} 
%     }
%     \subfigure[Real Author]{
%         \includegraphics[width=\columnwidth]{figure/lab2_2.png} 
%     }
%     \subfigure[World Fact]{
%         \includegraphics[width=\columnwidth]{figure/lab2_3.png} 
%     }
%     \caption{Impact of sample selection on unlearning evaluation: variability in performance across different LLM unlearning methods.}
%     \label{fig:three_images}
% \end{figure}

\paragraph{Measure the Unlearning Difficulty of Samples.} We argue that the primary cause of this bias lies in the varying direction and magnitude of parameter updates required to meet constraints when unlearning different samples. 
Specifically, even with the same unlearning algorithm, some samples are inherently harder to unlearn as they demand more frequent and larger parameter updates. 
This increases the complexity of the unlearning process and can negatively impact other model capabilities, leading to instability in unlearning performance. 
As a result, if the selected samples are easier to unlearn, the model's performance may appear significantly less damaged. 
However, this improvement stems from sample selection bias rather than enhancements in the unlearning algorithm itself.
Such bias can distort the evaluation of existing LLM unlearning algorithms, leading to misleading conclusions about their effectiveness. 
To address this, it is crucial to develop a metric that quantifies the unlearning difficulty of samples. 
This would enable a deeper understanding of LLM unlearning behavior and guide the development of more efficient and reliable methods for practical applications.

\subsection{Analyzing the Unlearning Difficulty of Sample} \label{met-uds}

To quantify the unlearning difficulty of a sample, a natural approach is to measure the change in model parameters before and after unlearning: $\Delta\boldsymbol{\theta}=\|\boldsymbol{\theta}^*-\boldsymbol{\theta}^{\prime}\|_2^2$, where $\boldsymbol{\theta}^*$ represents the parameters after unlearning.
However, as $\boldsymbol{\theta}^{\prime}$ is typically unknown in practice, this direct computation is infeasible.
One potential solution is to approximate this measure via bi-level optimization.
Yet, such methods~\cite{sekhari2021remember,thudi2022unrolling} often require second-order information (e.g., Hessian matrix inversion), leading to prohibitive computational costs for LLMs.
Thus, an alternative metric is needed to estimate unlearning difficulty effectively while minimizing computational overhead.

\paragraph{Definition of Unlearning Difficulty.} Inspired by neuroscience research~\cite{kim1992modality,squire1995retrograde,frankland2005organization,konrad2011long}, studies on human memory indicate that long-term memories (e.g., personal experiences or core skills) are generally robust to minor Traumatic Brain Injuries (mTBI), whereas short-term memories are more prone to disruption. 
This suggests that the brain exhibits varying difficulty levels when forgetting (i.e., unlearning) different types of knowledge.
Building on this analogy, we extend this finding to LLMs to assess the unlearning difficulty of specific samples. 
Similar to human memory, we hypothesize that samples with high unlearning difficulty (analogous to long-term memories) will exhibit minimal changes in the generated probability distribution under minor parameter perturbations (analogous to mTBI). 
In contrast, samples that are easier to unlearn will display more significant changes.

Specifically, we propose an initial metric, $\mathrm{MRD}$, to quantify unlearning difficulty, defined as:
\begin{equation}
   \mathrm{MRD}(\boldsymbol{x}^i;\boldsymbol{\theta}) = \left| \sum_{t=1}^{n_i} P_t(\boldsymbol{\theta}) - P_t(\boldsymbol{\theta}+\boldsymbol{\delta}) \right|,
   \label{eq:mrd_ori}
\end{equation}
where $P_t(\boldsymbol{\theta}) = \log p(x_t|\boldsymbol{x}_{<t};\boldsymbol{\theta})$ and $\boldsymbol{\delta}$ represents a small random perturbation applied to the model parameters. 
However, this preliminary metric has two key limitations:
\begin{enumerate}[leftmargin=*] \setlength{\itemsep}{2pt}
    %\item \textbf{Length Bias.} Longer samples yield lower generation probabilities, making $\mathrm{MRD}$ values artificially smaller, thereby introducing an unfair bias.
    \item \textbf{Limited Perturbation Scope.} Using a single perturbation direction may fail to capture the broader impact of parameter variations on the generation probability.
    \item \textbf{Absolute Metric Bias.} Absolute changes in probabilities may unfairly penalize samples with inherently low generation probabilities.
\end{enumerate}
To address these limitations, we propose improvements including sample length normalization, a global perturbation mechanism, and relative measures. 
The refined metric for unlearning difficulty is formally defined in Definition~\ref{de:mrd}.

\begin{definition}
    For an LLM with parameters $\boldsymbol{\theta}$, the difficulty of unlearning a sample $\boldsymbol{x}^i$ ($\mathrm{MRD}$) is defined as:
    \begin{equation}
    \small
        \mathrm{MRD}(\boldsymbol{x}^i;\boldsymbol{\theta}) = \left| \mathbb{E}_{\boldsymbol{\delta} \sim \mathcal{N}(0,\boldsymbol{\boldsymbol{\sigma}^2}I)} \sum_{t=1}^{n_i}  \left( \frac{P_t(\boldsymbol{\theta}) - P_t(\boldsymbol{\theta}+\boldsymbol{\delta})}{P_t(\boldsymbol{\theta})} \right) \right|,
    \normalsize
    \end{equation}
    where $\boldsymbol{\delta}$ is a Gaussian perturbation vector with mean 0 and variance $\boldsymbol{\boldsymbol{\sigma}^2}$.
    \label{de:mrd}
\end{definition}
A smaller $\mathrm{MRD}$ value indicates less fluctuation in the generation probability under parameter perturbations, implying higher unlearning difficulty. 
In contrast, a larger $\mathrm{MRD}$ suggests lower unlearning difficulty.

\begin{theorem}
    \textbf{Approximation of MRD.} Assuming that $P_t(\boldsymbol{\theta})$ and $P_t(\boldsymbol{\theta}+\boldsymbol{\delta})$ are non-zero, and $\boldsymbol{\delta} \sim \mathcal{N}(0,\boldsymbol{\boldsymbol{\sigma}^2}I)$ represents a small perturbation where $\boldsymbol{\boldsymbol{\sigma}^2}$ is sufficiently small, the $\mathrm{MRD}$ can be approximated as follows: 
    \begin{equation}
        \mathrm{MRD}(\boldsymbol{x}^i; \boldsymbol{\theta}) 
        \approx \frac{\boldsymbol{\boldsymbol{\sigma}^2}}{2} \sum_{t=1}^{n_i} \frac{\mathrm{Tr}(H_t)}{P_t(\boldsymbol{\theta})}
        = \frac{\boldsymbol{\boldsymbol{\sigma}^2}}{2}\sum_{t=1}^{n_i} \frac{\Delta P_t(\boldsymbol{\theta})}{P_t(\boldsymbol{\theta})},
        \label{eq:appro}
    \end{equation}
    where $H_t=\nabla^2P_t(\boldsymbol{\theta})$ represents the Hessian matrix of $P_t(\boldsymbol{\theta})$ w.r.t $\boldsymbol{\theta}$ and $\Delta P_t(\boldsymbol{\theta})$ denotes the Laplacian of $P_t(\boldsymbol{\theta})$.
    \label{the:appro}

    \begin{proof}
        The proof can be found in Appendix~\ref{sec: pro-app}.
    \end{proof}
\end{theorem}

\paragraph{Interpretation of MRD.} For the reasonableness of $\mathrm{MRD}$, Theorem~\ref{the:appro} shows that $\mathrm{MRD}(\boldsymbol{x}^i; \boldsymbol{\theta})$ is proportional to the Hessian trace $\mathrm{Tr}(H_t)$, which quantifies the second-order sensitivity (local curvature) of the $t$-th word's generation probability w.r.t. model parameters $\boldsymbol{\theta}$.
This implies that $\mathrm{MRD}$ effectively represents a weighted average of the model's local curvature.
In unlearning tasks, higher local curvature corresponds to lower model sensitivity to parameter changes, necessitating larger or more iterations of parameter updates to achieve the unlearning goal.
Thus, $\mathrm{MRD}$ serves as a reasonable metric for unlearning difficulty.

% \begin{theorem}
%     \textbf{Computational complexity of MRD.} For a sample $\boldsymbol{x}^i = \{x_1, \ldots, x_{n_i} \}$ using $K$ Monte Carlo samples to compute $\mathrm{MRD}$, where the number of model parameter is $d$, the computational complexity of $\mathrm{MRD}$ is $O(K\cdot n_i\cdot d)$.
%     \label{the:compute}

%     \begin{proof}
%         The proof can be found in Appendix~\ref{sec: pro-com}.
%     \end{proof}
% \end{theorem}

\begin{algorithm}[tb]
   \caption{Computation implementation of $\mathrm{MRD}$}
   \label{alg:compute_mrd}
\begin{algorithmic}
   \STATE {\bfseries Input:} Sample sequence $\boldsymbol{x}^i = \{x_t\}_{t=1}^{n_i}$ of length $n_i$; model parameters $\boldsymbol{\theta} \in \mathbb{R}^{d}$; disturbance variance $\boldsymbol{\boldsymbol{\sigma}^2}$; number of Monte Carlo samples $K$.
   \STATE {\bfseries Output:} The $\mathrm{MRD}$ value of sample $\boldsymbol{x}^i$.
   \STATE {\bfseries Initialize:} $\mathrm{MRD}_{\text{sum}} = 0$.
   \FOR{$k=1$ {\bfseries to} $K$}
      \STATE Sample disturbance vector $\boldsymbol{\delta}_k \in \mathbb{R}^d$ from $\mathcal{N}(0, \boldsymbol{\boldsymbol{\sigma}^2} I)$.
      \STATE Initialize $\Delta_{\text{sum}} = 0$.
      \FOR{$t=1$ {\bfseries to} $n_i$}
         \STATE Compute $P_t(\boldsymbol{\theta}) = \log p(x_t | x_{<t}; \boldsymbol{\theta})$.
         \STATE Compute $P_t(\boldsymbol{\theta} + \boldsymbol{\delta}_k) = \log p(x_t | x_{<t}; \boldsymbol{\theta} + \boldsymbol{\delta}_k)$.
         %\STATE Compute $\Delta_t = \frac{|P_t(\boldsymbol{\theta}) - P_t(\boldsymbol{\theta} + \delta_k)|}{P_t(\boldsymbol{\theta})}$.
         \STATE Compute $\Delta_t = \frac{P_t(\boldsymbol{\theta}) - P_t(\boldsymbol{\theta} + \boldsymbol{\delta}_k)}{P_t(\boldsymbol{\theta})}$.
         \STATE Update $\Delta_{\text{sum}} \gets \Delta_{\text{sum}} + \Delta_t$.
      \ENDFOR
      %\STATE Compute $\mathrm{MRD}_k = \frac{\Delta_{\text{sum}}}{n_i}$.
      \STATE Compute $\mathrm{MRD}_k = \left| \Delta_{\text{sum}} \right|$.
      \STATE Update $\mathrm{MRD}_{\text{sum}} \gets \mathrm{MRD}_{\text{sum}} + \mathrm{MRD}_k$.
   \ENDFOR
   \STATE Compute final $\mathrm{MRD}$ value: $\mathrm{MRD}(x^i; \boldsymbol{\theta}) = \frac{\mathrm{MRD}_{\text{sum}}}{K}$.
   \STATE \textbf{Return:} $\mathrm{MRD}(\boldsymbol{x}^i; \boldsymbol{\theta})$.
\end{algorithmic}
\end{algorithm}

\paragraph{Computational Complexity of MRD.} In practical implementation, the $\mathrm{MRD}$ quantifies the normalized variation in the generation probability of a sample $\boldsymbol{x}^i$ under parameter perturbations $\boldsymbol{\delta} \sim \mathcal{N}(0,\boldsymbol{\boldsymbol{\sigma}^2}I)$.
As the expectation cannot be computed analytically, it is approximated via Monte Carlo sampling.
Algorithm~\ref{alg:compute_mrd} outlines the procedure.
For a sample $\boldsymbol{x}^i = {x_1, \ldots, x_{n_i} }$ with $K$ Monte Carlo samples, the computational complexity of $\mathrm{MRD}$ is $\mathcal{O}(K \cdot n_i \cdot d)$, where $d$ is the number of model parameters. This demonstrates that $\mathrm{MRD}$ scales linearly with $d$, ensuring computational efficiency.

\paragraph{Characteristics Influencing MRD.} As stated in Theorem~\ref{the:appro}, %the approximate $\mathrm{MRD}$ value is given by: 
%
% $$
% \mathrm{MRD}(x^i;\boldsymbol{\theta}) 
% \approx \frac{\boldsymbol{\boldsymbol{\sigma}^2}}{2n_i}\sum_{t=1}^{n_i}\frac{\mathrm{Tr}(H_t)}{P_t(\boldsymbol{\theta})} 
% = \frac{\boldsymbol{\boldsymbol{\sigma}^2}}{2n_i}\sum_{t=1}^{n_i} \frac{\Delta P_t(\boldsymbol{\theta})}{P_t(\boldsymbol{\theta})},
% $$
% %
% where $\Delta P_t(\boldsymbol{\theta})$ denotes the Laplacian of the log-likelihood.
%
%This suggests that 
%
$\mathrm{MRD}$ is proportional to the local geometric curvature ($\Delta P_t(\boldsymbol{\theta})$) and inversely related to the normalization factor ($P_t(\boldsymbol{\theta})$), we conduct the following analysis: 

\begin{itemize}[leftmargin=*] \setlength{\itemsep}{2pt}
    \item For samples with smooth output distributions, such as syntactically simple and structurally clear ones (e.g., ``The cat is sleeping."), the local geometric curvature is relatively small (i.e., $\Delta(\log p(x_t|x_{<t};\boldsymbol{\theta}))$ is small). 
    Consequently, their $\mathrm{MRD}$ values are low, indicating higher resistance to unlearning. 
    In contrast, low-frequency samples from long-tail distributions or those with nested syntax and complex modifications (e.g., ``The intricacies of quantum mechanics perplex many scientists.") exhibit steeper distributions with sharper parameter-space variations. 
    These samples often have higher $\mathrm{MRD}$ values, making them more susceptible to perturbations and unlearning.
    \item If a sample's generation probability ($P_t(\boldsymbol{\theta})$) is high, its corresponding $\mathrm{MRD}$ will be small, indicating greater resistance to unlearning. 
    Intuitively, high-probability samples (e.g., ``I love reading books.") are often easier for the model to learn, as they frequently appear in the training set or share contextual similarities with other samples. 
    In contrast, samples with complex syntax or rare vocabulary (e.g., ``The sesquipedalian lecturer pontificated endlessly.") exhibit larger changes in generation probabilities under parameter perturbations, making them more susceptible to unlearning.
\end{itemize}

In Section~\ref{exp:res}, we validate these conclusions through extensive experiments, further confirming the effectiveness and reliability of the $\mathrm{MRD}$ metric in quantifying sample unlearning difficulty.

% \subsection{Factors Influencing the Unlearning Difficulty}

\subsection{MRD-based Weighted Sampling Method}

Building on $\mathrm{MRD}$, current LLM unlearning algorithms can be refined for greater effectiveness and efficiency. 
Drawing inspiration from curriculum learning, we propose a straightforward enhancement, i.e., weighted sampling. 
This approach ranks $\mathrm{MRD}$ values and adjusts sampling probabilities, prioritizing easily forgettable samples before harder ones, serving as a general, plug-and-play strategy compatible with current unlearning methods. 
For analytical clarity, we extend the commonly used Stochastic Gradient Ascent (SGA) method into a Curriculum Gradient Ascent (CGA) framework leveraging $\mathrm{MRD}$.

%
%In this paper, we modify SGA as a demonstration example.

%The update rate $R(U)$ measures the efficiency of an algorithm $U$ in performing unlearning, defined by the number of updates required to meet the unlearning constraint under a given computational complexity.
%
\begin{definition}
    For an unlearning algorithm $\mathcal{U}$, the unlearning efficiency is defined as $E(\mathcal{U}) = \frac{1}{M(\mathcal{U}) \cdot C(\mathcal{U})}$, where $M(\mathcal{U})$ is the number of updates needed to meet the unlearning goal, and $C(\mathcal{U})$ is the computational cost per update.
\end{definition}

\begin{definition}
    When the update magnitude per iteration is fixed, the average number of updates required to unlearn a sample $\boldsymbol{x}^i$ can be regarded as $I(\boldsymbol{x}^i) = {1} / {\mathrm{MRD(\boldsymbol{x}^i;\boldsymbol{\theta})}}$.
\end{definition}

\paragraph{Analyzing SGA.} For the algorithm $\mathcal{U}_{\text{SGA}}$, the procedure involves two steps:
(i) Randomly sample $\boldsymbol{x}^i \in \mathcal{D}_F$ at each iteration.
(ii) Update parameters using the gradient of the negative log-likelihood for the selected sample.
%
%Here, $\mathrm{MRD}(\boldsymbol{x}^i; \boldsymbol{\theta})$ represents the average number of updates needed to unlearn $\boldsymbol{x}^i$. 
%
Assuming uniform selection probability $p_i = 1 / {N_f}$, the total updates required for unlearning are:
$M(\mathcal{U}_{\text{SGA}}) = N_f \sum_{i=1}^{N_f} I(\boldsymbol{x}^i)$.
With per-update computational complexity $\mathcal{O}(d)$ and sampling complexity $\mathcal{O}(1)$, the unlearning efficiency is $E(\mathcal{U}_{\text{SGA}})= 1 / ({N_f \sum_{i=1}^{N_f} I(\boldsymbol{x}^i) \cdot \mathcal{O}(d)})$.
%
% \begin{equation}
%     E(\mathcal{U}_{\text{SGA}})=\frac{1}{N_f \sum_{i=1}^{N_f} \mathrm{MRD}(\boldsymbol{x}^i; \boldsymbol{\theta}) \cdot \mathcal{O}(d)}.
% \end{equation}

\paragraph{Analyzing CGA.} The $\mathrm{MRD}$-based method $U_{\text{CGA}}$ comprises three key steps, as outlined in Algorithm~\ref{alg:Curriculum_unlearning}:
(i) Compute $\mathrm{MRD}$ values for all samples.
(ii) Select samples based on $\mathrm{MRD}$, prioritizing those with lower unlearning difficulty.
(iii) Apply gradient ascent updates to the selected samples.
The selection probability of a sample $\boldsymbol{x}^i$ is defined as $p_i = I(\boldsymbol{x}^i) / {\sum_{j=1}^{N_f} I(\boldsymbol{x}^j)}$. 
This results in a total unlearning update cost of $M(\mathcal{U}_{\text{CGA}}) = \sum_{j=1}^{N_f} I(\boldsymbol{x}^j)$.
The complexity of $\mathcal{U}_{\text{CGA}}$ includes $\mathcal{O}(N_f \cdot d)$ for $\mathrm{MRD}$ computation and $\mathcal{O}(d)$ for parameter updates. 
Since $\mathrm{MRD}$ is recalculated every $m$ epochs, its overhead is minimal. 
Unlearning efficiency is $E(\mathcal{U}_{\text{CGA}}) = 1 / ({\sum_{j=1}^{N_f} I(\boldsymbol{x}^j) \cdot \mathcal{O}(d)})$.
%
% \begin{equation}
%     E(\mathcal{U}_{\text{CGA}}) = \frac{1}{\sum_{j=1}^{N_f} \mathrm{MRD}(\boldsymbol{x}^j; \boldsymbol{\theta}) \cdot \mathcal{O}(d)}.
% \end{equation}

The CGA method achieves a significantly higher unlearning efficiency than the SGA algorithm, with $E(\mathcal{U}_{\text{CGA}}) \approx N_f E(\mathcal{U}_{\text{SGA}})$. 
This advantage is more pronounced for large unlearning sets. 
Thus, under equivalent computational cost (e.g., a fixed number of updates), $\mathcal{U}_{\text{CGA}}$ demonstrates superior unlearning performance, reducing the gap between the model's unlearning completeness and the target threshold while preserving other capabilities.
The comparison of improvements for other LLM unlearning methods will be discussed in subsequent experiments.

\section{Experiments}

\subsection{Experiment Setups}

\paragraph{Unlearning Tasks and Datasets.} To validate the $\mathrm{MRD}$ metric and $\mathrm{MRD}$-enhanced methods, we follow experimental setups from prior work~\cite{jia2024wagle} and evaluate across four mainstream LLM unlearning datasets and tasks:
\begin{itemize}[leftmargin=*] %\setlength{\itemsep}{-\itemsep}
\setlength{\itemsep}{2pt}
    \item \textbf{TOFU}~\cite{maini2024tofu}, virtual author information unlearning.
    \item \textbf{WMDP}~\cite{li2024wmdp}, unlearning harmful capabilities. 
    \item \textbf{WHP}~\cite{eldan2023s}, copyright information removal. 
    \item \textbf{SAFE}~\cite{ji2024beavertails}, unlearning model toxic responses. 
\end{itemize}
% i) , 
% %
% ii) , 
% %
% iii) , 
% %
% and iv) . 
%
Detailed dataset information can be found in Appendix~\ref{app:data-con}.

\paragraph{Models.} For the TOFU task, we follow the original setup and utilize the LLaMA2-7B-chat~\cite{touvron2023llama}. 
For the WMDP task, we employ the Zephyr-7B-beta~\cite{tunstall2023zephyr}, consistent with its benchmark. 
In the WHP task, we perform LoRA fine-tuning on the LLaMA2-7B~\cite{touvron2023llama} using the complete Harry Potter series. 
Finally, for the validation of the 
SAFE dataset, we conduct experiments using the LLaMA2-7B.

\paragraph{Evaluation Metrics.}

We assess unlearned LLM performance through two dimensions: Unlearning Completeness (\textbf{UC}) and Model Utility (\textbf{UT}). 
UC quantifies the model's ability to unlearn targeted data, while UT evaluates the impact of unlearning on unrelated tasks.
%
%The evaluation metrics are summarized below.
%
For the TOFU task, UC is measured using three metrics: Unlearning Accuracy (UA), Membership Inference Attack (MIA), and Rouge-L Recall (RR). 
UA is represented as 1-Forget Accuracy (FA)~\cite{jia2024wagle}, where FA measures the model’s accuracy on the forget set, with higher UA indicating better unlearning completeness.
MIA evaluates the area under the ROC curve (AUC) using the Min\text{-}$k\%$ Prob~\cite{shi2023detecting} method to detect training set membership. Higher MIA scores suggest improved model confidence in unlearning.
RR=1-Rouge-L is used for averaged evaluations, where Rouge-L is also measured over the forget set, with higher RR scores indicating better performance.
UT is assessed via accuracy and Rouge-L recall on the retain set.
For WMDP, UC is evaluated using 1-FA on WMDP-Bio and WMDP-Cyber subsets, with UT measured by zero-shot accuracy on the MMLU dataset~\cite{hendrycks2020measuring}.
For WHP, UC is determined using Rouge-L on 300-token completions from Harry Potter-based instructions, while UT is evaluated through Perplexity (PPL) on Wikitext~\cite{merity2016pointer} and averaged zero-shot accuracy across tasks via the Language Model Evaluation Harness~\cite{gao2021framework}.
For SAFE, UC is assessed using Toxic-BERT~\cite{hanu2020detoxify} scores on toxic prompts from the SAFE test set, with UT evaluation mirroring that of WHP.
Detailed descriptions can be found in Appendix~\ref{app:eva-con}.

\paragraph{Baselines.} We assess the $\mathrm{MRD}$ metric's efficacy on mainstream unlearning baselines, including gradient-based methods (GA~\cite{jang2023knowledge} and GradDiff~\cite{yao2024large}) and preference optimization methods (PO~\cite{maini2024tofu} and NPO~\cite{zhang2024negative}).
For each baseline, we propose an $\mathrm{MRD}$-weighted sampling strategy to refine the unlearning sequence, yielding an $\mathrm{MRD}$-enhanced method.
Comparative analysis is conducted against original baselines, with results averaged over five independent trials.

\paragraph{Training Setup.} We set the AdamW~\cite{loshchilov2017decoupled} optimizer as the default optimization algorithm, with a learning rate of $5e-5$. 
The perturbation intensity $\boldsymbol{\sigma}$ is set to $1e-5$, and the number of Monte Carlo sampling iterations $K$ for calculating $\mathrm{MRD}$ is set to 200.
%
%More implementation details can be found in Appendix~\ref{app:unlearn-con}.
%
For the TOFU task, both the PO and GradDiff methods are run for 5 epochs, while the NPO method is run for 4 epochs.
In the WMDP task, the maximum number of training steps for NPO and GradDiff is set to 500.
For the WHP and SAFE tasks, 5 epochs are conducted.

\subsection{Experiment Results} \label{exp:res}

\begin{figure}[htbp]
    \centering
    \subfigure[GA]{
        \includegraphics[width=0.3\linewidth]{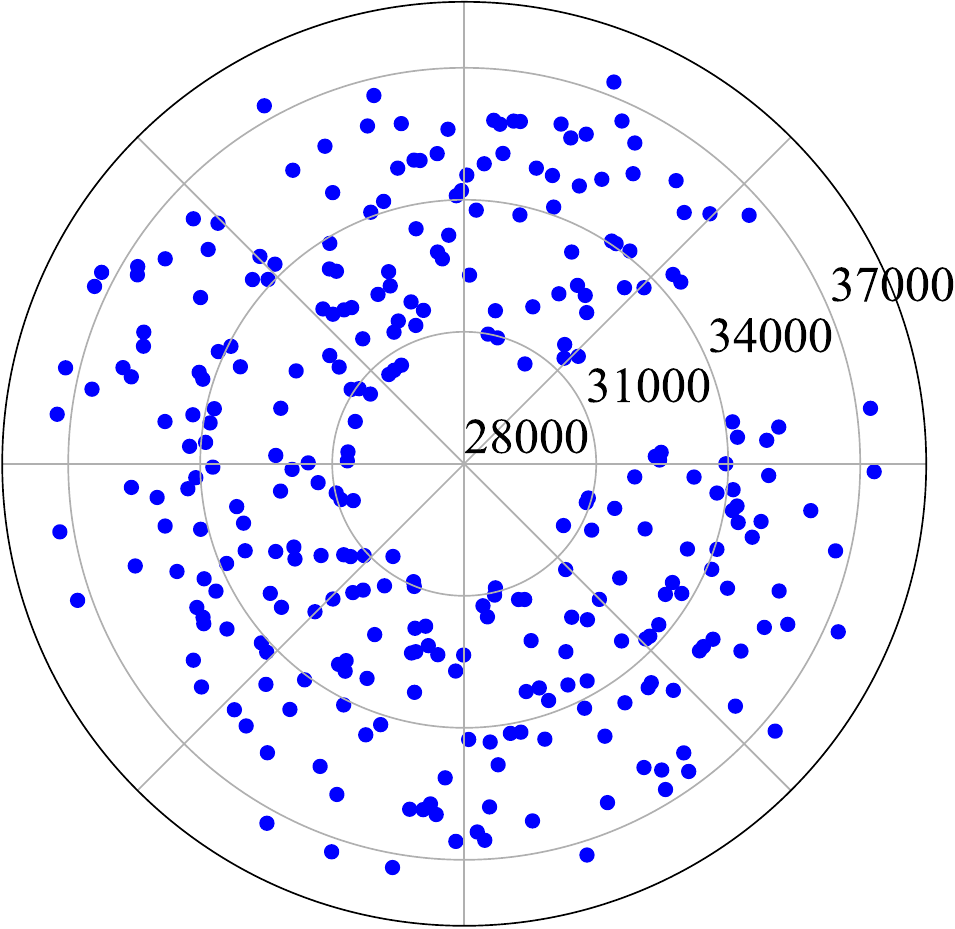} 
    }
    \subfigure[GradDiff]{
        \includegraphics[width=0.3\linewidth]{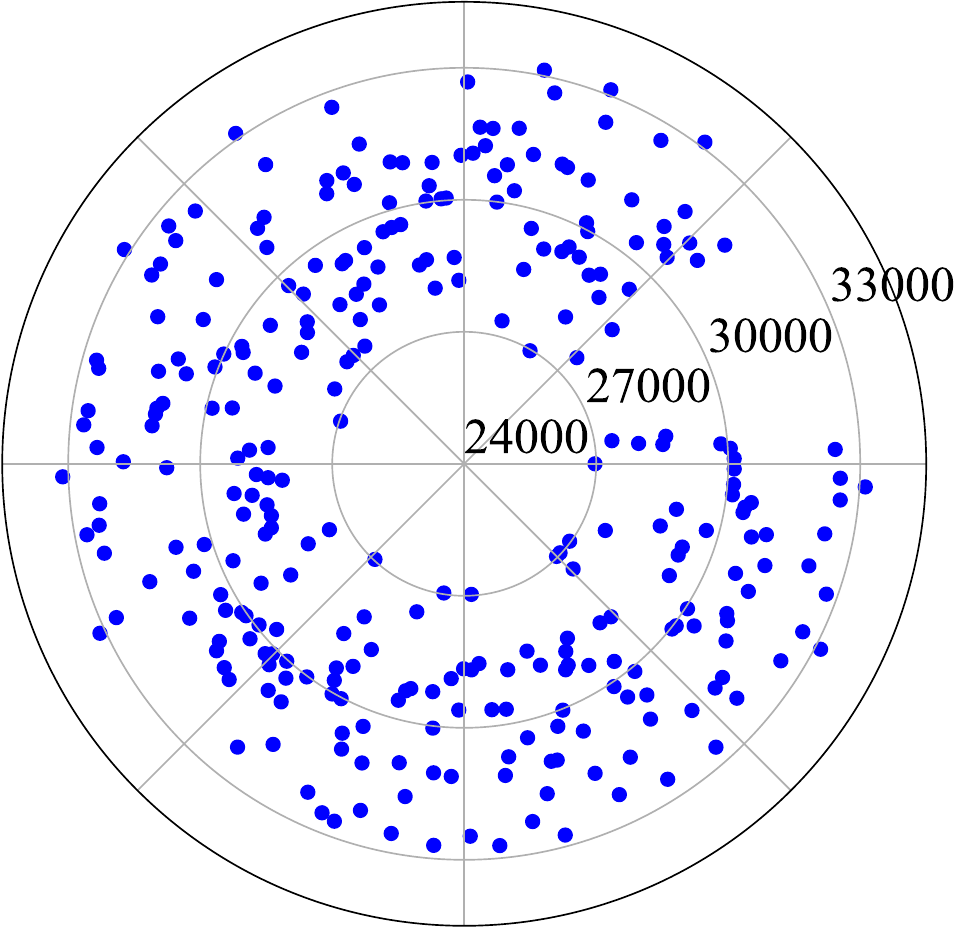} 
    }
    \subfigure[NPO]{
        \includegraphics[width=0.3\linewidth]{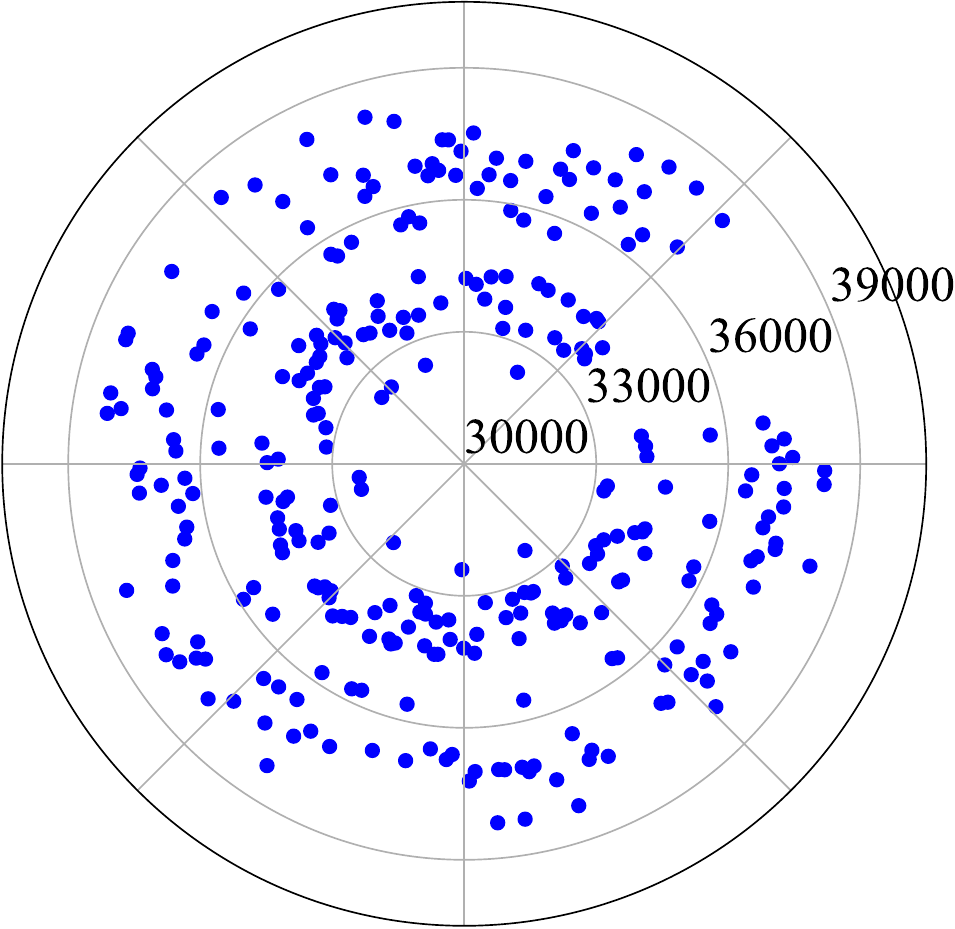} 
    }
    \caption{Comparison of unlearning difficulty across different sample sets in GA, GradDiff, and NPO. In these polar coordinates, samples are uniformly distributed in terms of angle, while the distance denotes the average absolute value of parameter changes.}
    \label{fig:diff-ud}
\end{figure}

\paragraph{Differences in Unlearning Difficulty.} We confirm that the magnitude of model parameter changes during the unlearning of different samples in the TOFU task exhibits notable variability, indicating non-uniform unlearning difficulty across samples.
Since parameter changes from unlearning a single sample are typically small, we employ a sample concatenation strategy to amplify the analysis.
Specifically, 40 samples are randomly selected with replacements from the unlearning set and concatenated into composite samples, resulting in 300 such samples.
For each composite sample, unlearning is performed using an existing LLM unlearning baseline with an early stopping condition (Appendix~\ref{app:con-ear}).
We then compute the average absolute value of parameter changes post-unlearning to assess the impact of different samples.
As shown in Figure~\ref{fig:diff-ud}, the results demonstrate significant variability in parameter changes across samples.%, with inconsistent update directions.
This confirms that unlearning difficulty differs among samples, and the choice of unlearning samples substantially influences unlearning performance.

\begin{figure}[htbp]
    \centering
    \subfigure[TOFU]{
        \includegraphics[width=0.47\linewidth]{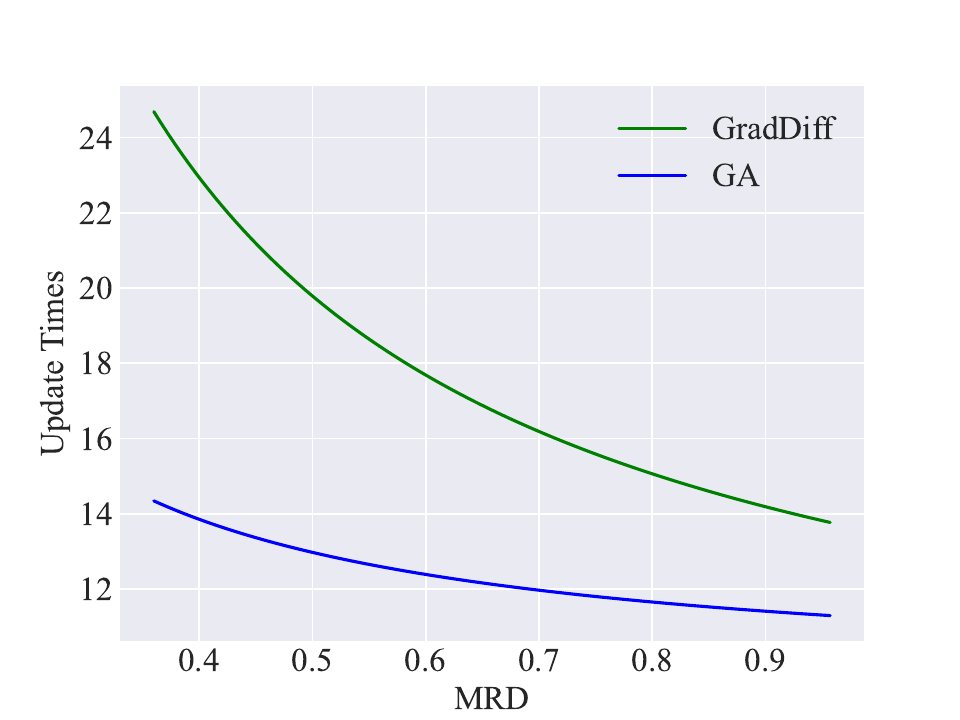} 
    }
    \subfigure[WMDP]{
        \includegraphics[width=0.47\linewidth]{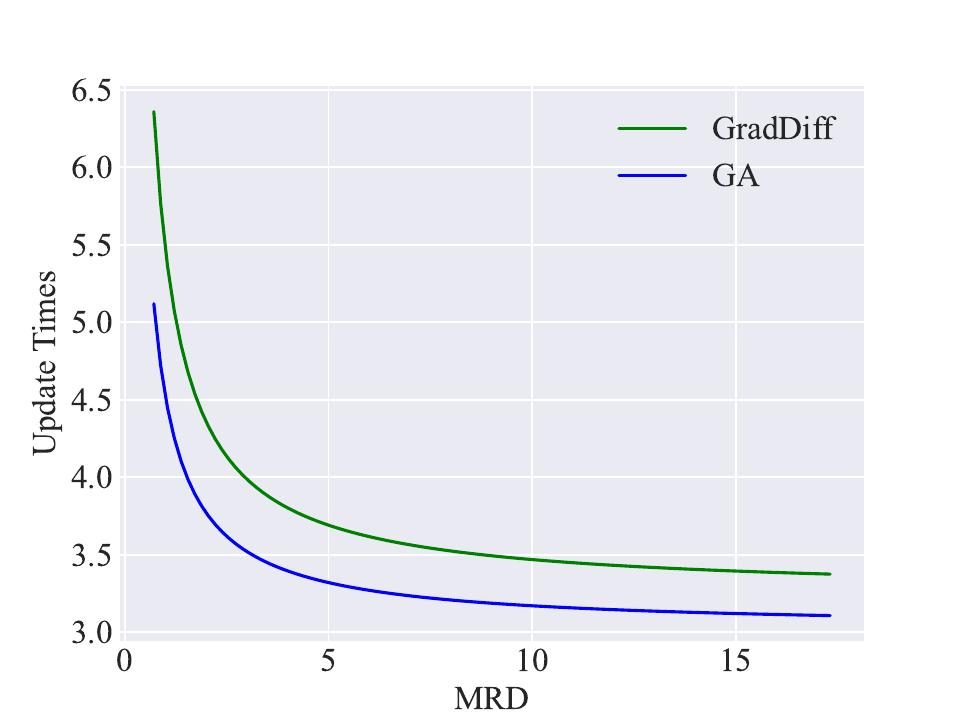} 
    }
    \caption{The relationship between the MRD value and the number of unlearning updates (i.e., unlearning difficulty).}
    \label{fig:effect-mrd}
\end{figure}

\paragraph{Effectiveness of MRD.} To validate the effectiveness of our proposed $\mathrm{MRD}$ metric, we conduct experiments on two tasks: TOFU and WMDP.
For each task, 10 samples are randomly selected, and their $\mathrm{MRD}$ values are computed.
To further evaluate the metric's utility, we apply various LLM unlearning baselines to unlearn each sample.
Using identical hyperparameter settings, parameter update magnitudes, and early stopping conditions, we compare the number of updates required for unlearning across samples.
The experiment is repeated three times, with results shown in Figure~\ref{fig:effect-mrd}.
From it, we observe that $\mathrm{MRD}$ values effectively capture sample difficulty, aligning consistently with the required update counts for the same unlearning algorithm.
Moreover, the ranking of update counts across different methods remains generally consistent, suggesting that variability in unlearning behavior is an intrinsic property of the samples.

\paragraph{Characteristics Influencing MRD.} To explore characteristics influencing $\mathrm{MRD}$, enhance its interpretability, and guide future unlearning research, we conduct experiments on the TOFU task.
The unlearning sample set is categorized based on four criteria: semantic complexity, occurrence frequency, initial generation probability, and presence of rare words.
%
% Semantic complexity is quantified using XXX, with samples meeting the XX condition labeled as high-complexity.
Semantic complexity is quantified using lexical diversity indices and syntactic complexity measures\cite{jiang2021measuring}, with samples meeting the threshold of upper quartile values labeled as high-complexity.
Occurrence frequency is classified relative to the training set average, with high-frequency samples exceeding this threshold.
Initial generation probability is similarly categorized using the average probability as the threshold.
For rare words, a predefined high-frequency vocabulary serves as the baseline\cite{luong2015addressing}, and samples containing more than three occurrences of words outside this vocabulary are identified as rare-word samples.
% For rare words, XXX serves as the baseline, and samples containing more than three occurrences of words from XXX are identified as rare-word samples.

%
From the categorized set, 40 samples are randomly selected, and their $\mathrm{MRD}$ values are computed (Table~\ref{tab:attributes_mrd} in Appendix~\ref{app:cha-exp}).
Results reveal that high-frequency samples and those with high initial generation probabilities exhibit lower $\mathrm{MRD}$ values, indicating greater resistance to unlearning.
In contrast, high-complexity samples and those with rare words show higher $\mathrm{MRD}$ values, suggesting greater susceptibility to unlearning.
These findings align with the analysis in Section~\ref{met-uds}.

\begin{table*}[h!]
\centering
\footnotesize
\caption{Comparison of the $\mathrm{MRD}$-based weighted sampling method and the current unlearning baseline methods on TOFU. For the same baseline before and after improvement, we ensure consistent experimental settings. The optimal results are highlighted in \textbf{bold}.}
\label{tab:unlearning_performance}
\resizebox{\linewidth}{!}{
\begin{tabular}{cccccccccccc}
\toprule
\multirow{3}{*}{\textbf{Method}} & \multicolumn{4}{c}{\textbf{Unlearning Completeness (UC)}} & \multicolumn{7}{c}{\textbf{Model Utility (UT)}} 
\\
\cmidrule(lr){2-5} \cmidrule(lr){6-12}
& \multirow{2}{*}{UA ($\uparrow$)} & \multirow{2}{*}{MIA ($\uparrow$)} & \multirow{2}{*}{RR ($\uparrow$)}
&\multirow{2}{*}{Avg. ($\uparrow$)}&\multicolumn{2}{c}{Retain Set}&\multicolumn{2}{c}{Real Author}&\multicolumn{2}{c}{World Fact}&\multirow{2}{*}{Avg. ($\uparrow$)}
\\
& & & & & Acc. ($\uparrow$)&RR ($\uparrow$)& Acc. ($\uparrow$)&RR ($\uparrow$)& Acc. ($\uparrow$) &RR ($\uparrow$)&\\
\midrule
Original & 0.1475 & 0.4515 & 0.0204 & 0.2447 & 0.8575 & 0.9825 & 0.8900 & 0.9330 & 0.8632 & 0.8960 & 0.9037 \\
\midrule
SGA & 0.3725 & 0.4490 & 0.5722 & 0.4645 & 0.6125 & 0.4212 & 0.3512 & 0.3908 & 0.7094 & 0.7841 & 0.5449 \\
GradDiff & 0.8475 & 0.9977 & 0.9950 & 0.9467 & 0.7251 & 0.5131 & 0.7126 & 0.7473 & 0.8119 & 0.8547 & 0.7274 \\
PO & 0.7268 & 0.6478 & 0.9314 & 0.7686 & 0.6113 & 0.4190 & 0.6113 & 0.6988 & 0.7348 & 0.7862 & 0.6435 \\
NPO & 0.8354 & 0.9913 & 0.9821 & 0.9359 & 0.7432 & 0.5356 & 0.8269 & 0.8313 & 0.8262 & 0.8746 & 0.7729 \\

\midrule
CGA & 0.3825 & 0.4594 & 0.5781 & 0.4733 & 0.6575 & 0.4296 & 0.5147 & 0.5375 & 0.7436 & 0.7984 & 0.6135 \\
GradDiff + $\mathrm{MRD}$ & 0.8425 & \textbf{0.9997} & \textbf{0.9984} & \textbf{0.9469} & 0.7350 & 0.5253 & 0.7316 & 0.7321 & 0.8205 & 0.8561 & 0.7334 \\
PO + $\mathrm{MRD}$ & 0.7575 & 0.6512 & 0.9773 & 0.7953 & 0.6250 & 0.4216 & 0.6352 & 0.6963 & 0.7435 & 0.7792 & 0.6501 \\
NPO + $\mathrm{MRD}$ & \textbf{0.8525} & 0.9992 & 0.9854 & 0.9457 & \textbf{0.7775} & \textbf{0.5506} & \textbf{0.8913} & \textbf{0.8547} & \textbf{0.8462} & \textbf{0.8832} & \textbf{0.8005} \\

\bottomrule
\end{tabular}
\label{tab:main-tofu}}
\end{table*}

\paragraph{Effectiveness of MRD-based Weighted Sampling.} To evaluate the $\mathrm{MRD}$-based weighted sampling method (i.e., $\mathrm{MRD}$-enhanced method), we conduct experiments on four mainstream LLM unlearning tasks, comparing its performance with baseline methods regarding unlearning effectiveness and efficiency.
%
%Experimental details are in Appendix~\ref{app:hyp-set}. 
%
For the TOFU task, Table~\ref{tab:main-tofu} shows that the $\mathrm{MRD}$-enhanced method improves unlearning completeness by $1.12\%$ on average with the same number of update iterations.
$\mathrm{MRD}$ also boosts model utility, with an average gain of $2.72\%$, and achieves higher efficiency under equivalent early stopping conditions (i.e., meeting unlearning constraints).
These results validate our hypothesis that utilizing $\mathrm{MRD}$ to adjust the unlearning sequence can further optimize the performance of existing unlearning algorithms.
%
%confirm that optimizing the unlearning sequence enhances the performance of existing unlearning algorithms. 
%
Results for other tasks are reported in Appendix~\ref{app:eff-exp}.

\begin{figure}[htbp]
    \centering
    \subfigure[$\boldsymbol{\delta}$ perturbation]{\label{fig:para_1}
        \includegraphics[width=0.47\linewidth]{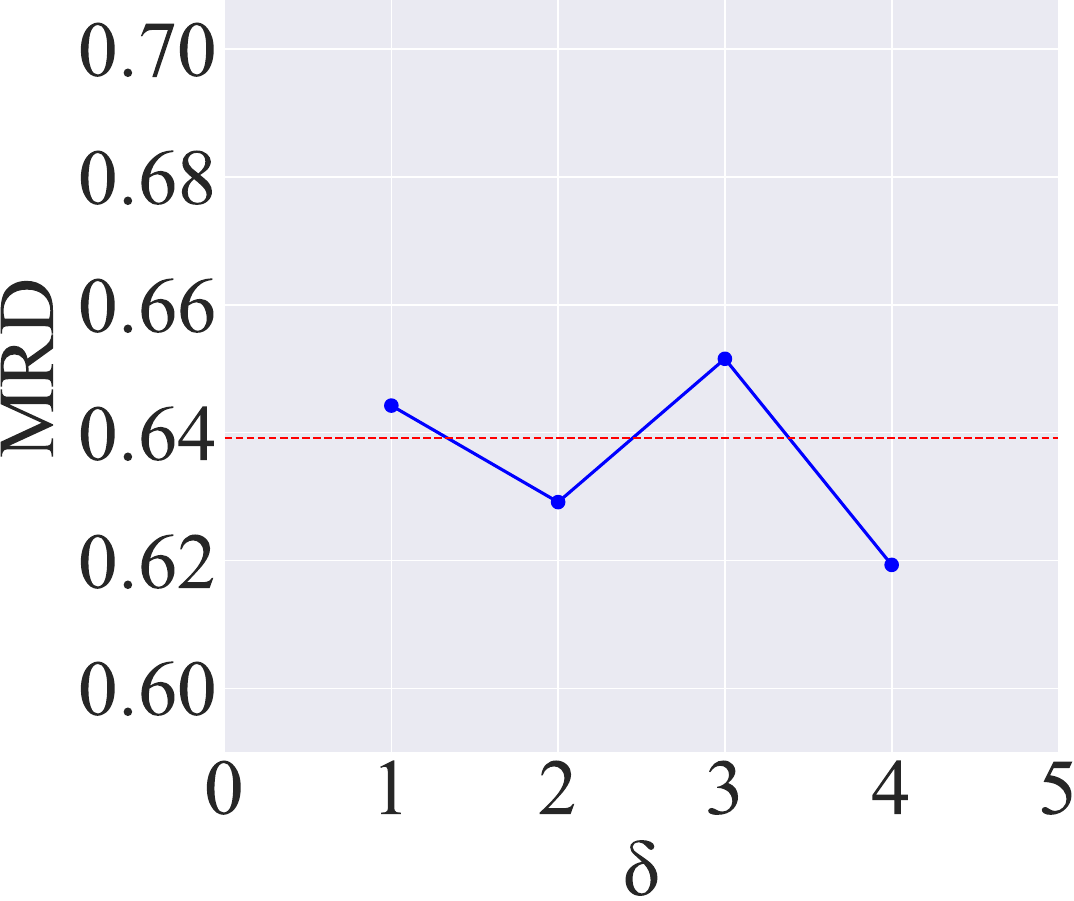} 
    }
    \subfigure[Monte Carlo $K$]{\label{fig:para_2}
        \includegraphics[width=0.47\linewidth]{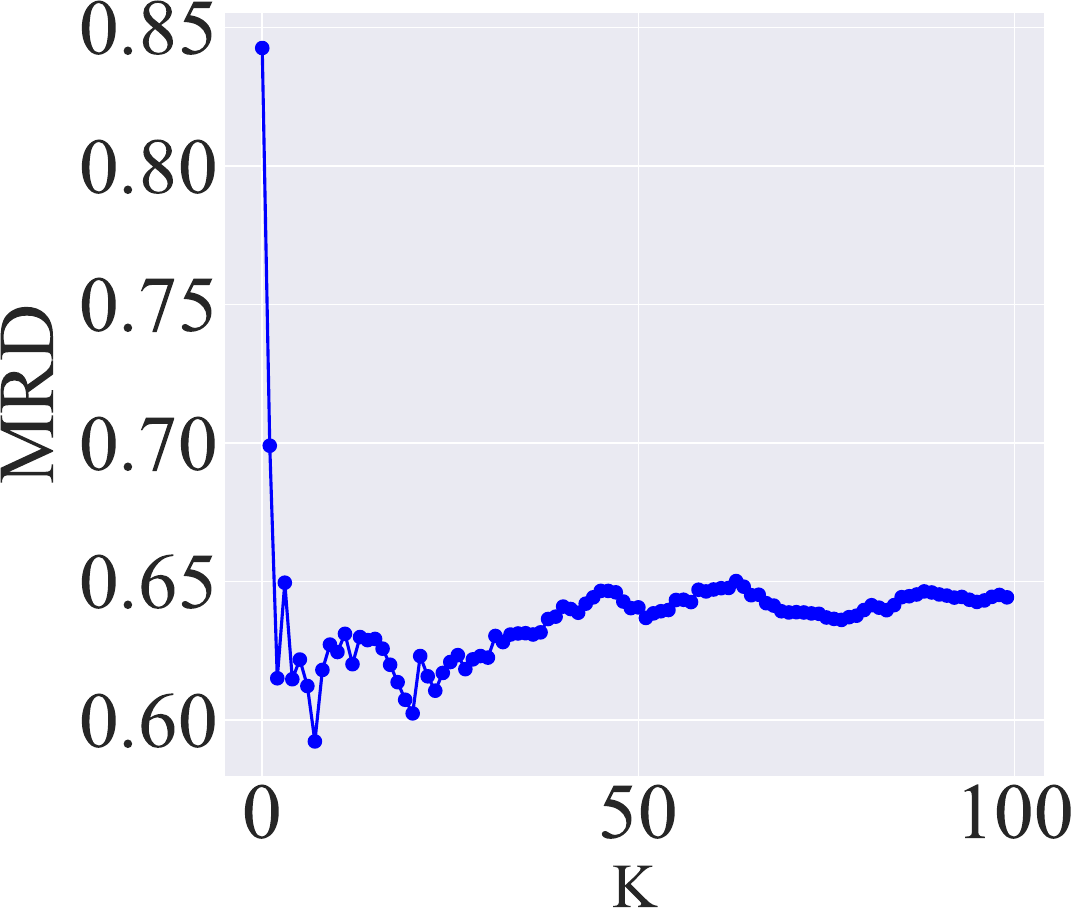} 
    }
    \caption{Parameter sensitivity of $\mathrm{MRD}$. (a) Effect of perturbation parameter $\boldsymbol{\delta}$, fluctuating around 0.64. (b) Effect of Monte Carlo samples $K$, with stability achieved at $K = 100$.}
    \label{fig:para}
\end{figure}

\paragraph{Parameter Sensitivity.} To evaluate the impact of the perturbation parameter $\boldsymbol{\delta}$ and the number of Monte Carlo samples $K$ on $\mathrm{MRD}$ calculation, we conduct experiments on the TOFU task.
Regarding the impact of $\boldsymbol{\delta}$ on the $\mathrm{MRD}$ calculation, we randomly select 20 samples, fix $K=100$, and compute $\mathrm{MRD}$ values with $\boldsymbol{\delta} \in \{1,2,3,4\}$, as shown in Figure~\ref{fig:para_1}.
Results indicate that as the value of $\boldsymbol{\delta}$ increases, the $\mathrm{MRD}$ value fluctuates around 0.64, suggesting that the calculation of $\mathrm{MRD}$ is not particularly sensitive to the choice of $\boldsymbol{\delta}$.
For computational simplicity, we choose $\boldsymbol{\delta} = 1$ in this paper.
Next, with $\boldsymbol{\delta} = 1$ fixed, we vary $K$ from 1 to 100 and compute the corresponding $\mathrm{MRD}$ values. 
Figure~\ref{fig:para_2} illustrates the variation of $\mathrm{MRD}$ values as $K$ increases. 
It can be observed that when $K$ is relatively small, the $\mathrm{MRD}$ calculation fluctuates significantly. 
However, as $K$ reaches 50, the $\mathrm{MRD}$ calculation gradually stabilizes, achieving optimal performance at $K = 100$.

% \begin{figure}[t]
% \vskip 0.2in
% \begin{center}
% \centerline{\includegraphics[width=\columnwidth]{figure/lab6_1.png}}
% \caption{MRD vs. perturbation parameter $\boldsymbol{\delta}$, showing optimal performance at $\boldsymbol{\delta} = 1$. }
% \end{center}
% \vskip -0.2in
% \end{figure}

% \begin{figure}[t]
% \vskip 0.2in
% \begin{center}
% \centerline{\includegraphics[width=\columnwidth]{figure/lab6_2.png}}
% \caption{MRD vs. Monte Carlo samples K, with stability achieved at K = 100.}
% \end{center}
% \vskip -0.2in
% \end{figure}
\section{Conclusion}

To improve the evaluation of existing LLM unlearning methods, we introduce a novel perspective by examining the unlearning characteristics of samples.
Inspired by neuroscience, we propose a metric, $\mathrm{MRD}$, to quantify the unlearning difficulty of samples.
Defined as the expected change in sample generation probability after applying Gaussian perturbations to model parameters, $\mathrm{MRD}$ demonstrates that unlearning difficulty varies significantly across samples, emphasizing the importance of sample selection in unlearning performance.
We further analyze the factors influencing the $\mathrm{MRD}$ value of samples, specifically identifying the characteristics of samples that make them harder or easier to unlearn. %, such as semantic complexity, sentence repetition frequency, and so on.
Then, we leverage these insights to propose an $\mathrm{MRD}$-based weighted sampling approach.
%We analyze $\mathrm{MRD}$ to identify common characteristics of samples that are harder or easier to unlearn and leverage these insights to propose an $\mathrm{MRD}$-based weighted sampling approach.
%
This approach refines existing unlearning methods by prioritizing the removal of easier-to-unlearn samples, improving both efficiency and effectiveness.
Extensive experiments confirm that incorporating sample-level characteristics, such as unlearning difficulty, enhances LLM unlearning methods.
Our analysis shows that $\mathrm{MRD}$ is not only reasonable and effective but also provides new directions and insights for subsequent studies on LLM unlearning. 
For instance, researchers could use $\mathrm{MRD}$ to reassess the rationality of LLM unlearning evaluation or improve existing methods based on $\mathrm{MRD}$, such as sample weighting.
%
%Our analysis demonstrates that $\mathrm{MRD}$ is both effective and insightful, offering new avenues for research, such as guiding unlearning evaluations or refining algorithms via sample weighting.
%
In summary, our work provides a fresh perspective on LLM unlearning, advancing the understanding of unlearning dynamics and improving method design.

%\section{Limitation}

% \section*{Acknowledgements}

% \textbf{Do not} include acknowledgements in the initial version of
% the paper submitted for blind review.

% If a paper is accepted, the final camera-ready version can (and
% usually should) include acknowledgements.  Such acknowledgements
% should be placed at the end of the section, in an unnumbered section
% that does not count towards the paper page limit. Typically, this will 
% include thanks to reviewers who gave useful comments, to colleagues 
% who contributed to the ideas, and to funding agencies and corporate 
% sponsors that provided financial support.

\nocite{langley00}

%%%%%%%%%%%%%%%%%%%%%%%%%%%%%%%%%%%%%%%%%%%%%%%%%%%%%%%%%%%%%%%%%%%%%%%%%%%%%%%
%%%%%%%%%%%%%%%%%%%%%%%%%%%%%%%%%%%%%%%%%%%%%%%%%%%%%%%%%%%%%%%%%%%%%%%%%%%%%%%
% APPENDIX
%%%%%%%%%%%%%%%%%%%%%%%%%%%%%%%%%%%%%%%%%%%%%%%%%%%%%%%%%%%%%%%%%%%%%%%%%%%%%%%
%%%%%%%%%%%%%%%%%%%%%%%%%%%%%%%%%%%%%%%%%%%%%%%%%%%%%%%%%%%%%%%%%%%%%%%%%%%%%%%
\newpage
\appendix
\onecolumn

\section{Proof of Theorem~\ref{the:appro}} \label{sec: pro-app}

The $\mathrm{MRD}$ metric is defined as:
$$
\mathrm{MRD}(\boldsymbol{x}^i;\theta) = \left| \mathbb{E}_{\boldsymbol{\delta} \sim \mathcal{N}(0,\boldsymbol{\sigma^2}I)} \sum_{t=1}^{n_i}  \left( \frac{P_t(\theta) - P_t(\theta+\boldsymbol{\delta})}{P_t(\theta)} \right) \right|,
$$
where $P_t(\theta)=\log p(x_t|x_{<t};\theta)$ represents the log-likelihood of the $t$-th token, $\boldsymbol{\delta} \sim \mathcal{N}(0,\boldsymbol{\sigma^2}I)$ is the parameter perturbation, and $n_i$ is the length of the sentence $\boldsymbol{x}^i$. The goal is to derive the relationship between $\mathrm{MRD}$ and the Hessian matrix.

To proceed, we perform a multivariate Taylor expansion of $P_t(\theta+\boldsymbol{\delta})$ up to the second-order term: 
$$
P_t(\theta+\boldsymbol{\delta})\approx P_t(\theta) + \nabla P_t(\theta)^\top \boldsymbol{\delta} + \frac12 \boldsymbol{\delta}^\top H_t \boldsymbol{\delta},
$$
where $\nabla P_t(\theta)$ is the gradient of $P_t(\theta)$ w.r.t $\theta$, and $H_t=\nabla^2P_t(\theta)$ is the Hessian matrix of $P_t(\theta)$ w.r.t. $\theta$. Substituting this expansion into $P_t(\theta)-P_t(\theta + \boldsymbol{\delta})$, we get:
$$
P_t(\theta) - P_t(\theta+\boldsymbol{\delta}) \approx - \nabla P_t(\theta)^\top \boldsymbol{\delta} - \frac12 \boldsymbol{\delta}^\top H_t \boldsymbol{\delta}.
$$
The relative change can then be expressed as:
$$
\frac{P_t(\theta)-P_t(\theta+\boldsymbol{\delta})}{P_t(\theta)} \approx -\frac{\nabla P_t(\theta)^\top \boldsymbol{\delta}}{P_t(\theta)}-\frac12 \frac{\boldsymbol{\boldsymbol{\delta}}^\top H_t\delta}{P_t(\theta)}.
$$
Substituting this expression into the $\mathrm{MRD}$ formula and averaging over all tokens in the sentence, we have:
$$
\mathrm{MRD}(\boldsymbol{x}^i;\theta) \approx \left| \mathbb{E}_{\boldsymbol{\delta} \sim \mathcal{N}(0,\boldsymbol{\sigma^2}I)} \sum_{t=1}^{n_i} \left( -\frac{\nabla P_t(\theta)^\top \boldsymbol{\delta}}{P_t(\theta)} - \frac{1}{2} \frac{\boldsymbol{\delta}^\top H_t \boldsymbol{\delta}}{P_t(\theta)} \right) \right|.
$$

Given that $\boldsymbol{\delta} \sim \mathcal{N}(0,\boldsymbol{\sigma^2}I)$, the expectation of $\boldsymbol{\delta}$ is $\mathbb{E}[\boldsymbol{\delta}]=0$. Consequently, the expectation of the first-order term vanishes: 
$$
\mathbb{E}\left[-\frac{\nabla P_t(\theta)^\top \boldsymbol{\delta}}{P_t(\theta)}\right]=0.
$$
For the second-order term, we compute the expectation using the properties of the multivariate normal distribution. Specifically, for $\boldsymbol{\delta} \sim \mathcal{N}(0,\boldsymbol{\sigma^2}I)$, the expectation of the quadratic form is: $\mathbb{E}[\boldsymbol{\delta}^\top H_t \boldsymbol{\delta} ]= \boldsymbol{\sigma^2} \operatorname{Tr}(H_t)$, where $\operatorname{Tr}(H_t)$ denotes the trace of the Hessian matrix $H_t$. Thus, the expectation of the second-order term becomes: 
$$
\mathbb{E} \left[ -\frac12\frac{\boldsymbol{\delta}^\top H_t \boldsymbol{\boldsymbol{\delta}}}{P_t(\theta)}\right]=-\frac{\boldsymbol{\sigma^2}}{2P_t(\theta)}\operatorname{Tr}(H_t).
$$
Since the expectation of the first-order term is zero, only the effect of the absolute value of the second-order term on the overall result needs to be considered. 
For the second-order term $-\frac{1}{2} \frac{\boldsymbol{\sigma^2} \mathrm{Tr}(H_t)}{P_t(\theta)}$, as $P_t(\theta)$ is always positive and the trace of the Hessian is typically positive, its sign is fixed and usually negative. 
Therefore, taking the absolute value only changes the sign but does not affect the overall value.
In this case, the absolute value of the expectation can be approximated by directly taking the absolute value of the second-order term. 
Consequently, the approximate expression for $\mathrm{MRD}$ is given as follows:
$$
\mathrm{MRD}(x^i;\theta)
\approx \frac{\boldsymbol{\sigma^2}}{2} \sum_{t=1}^{n_i} \frac{\operatorname{Tr}(H_t)}{P_t(\theta)} 
= \frac{\boldsymbol{\sigma^2}}{2}\sum_{t=1}^{n_i} \frac{\Delta P_t(\theta)}{P_t(\theta)}.
$$

% \subsection{Proof of Theorem~\ref{the:compute}} \label{sec: pro-com}

% Specifically, this process can be divided into two parts: perturbation sampling and generation probability computation.
% %
% \begin{enumerate}
%     \item \textbf{Perturbation sampling.} A total of $K$ samples are drawn, with a computational complexity of $O(K\cdot d)$.
%     \item \textbf{Generation probability computation.} For each perturbation, generation probabilities are computed for $n_i$ time steps, with a complexity of $O(n_i\cdot d)$. 
% \end{enumerate}
% %
% With $K$ total perturbations, the overall complexity for this part is $O(K\cdot n_i\cdot d)$.

\section{Algorithm of Curriculum Gradient Ascent Unlearning}

We present the algorithm of Curriculum Gradient Ascent Unlearning in Algorithm~\ref{alg:Curriculum_unlearning}.

\begin{algorithm}[tb]
   \caption{Curriculum Gradient Ascent Unlearning}
   \label{alg:Curriculum_unlearning}
\begin{algorithmic}[1]
   \STATE {\bfseries Input:} Model parameters $\boldsymbol{\theta} \in \mathbb{R}^{d}$; the forget set $\mathcal{D}_F = \{\boldsymbol{x}^1, \dots, \boldsymbol{x}^n\}$; unlearning difficulty metric function $\mathrm{MRD}(\boldsymbol{x}; \boldsymbol{\theta})$; $\mathrm{MRD}$ update interval $m$.
   \STATE {\bfseries Output:} Updated model parameter $\boldsymbol{\theta}$.
   \STATE {\bfseries Initialize:} Compute $\mathrm{MRD}(\boldsymbol{x}^i;\boldsymbol{\theta})$ for each sample $\boldsymbol{x}_i$, $i = 1, 2, \dots, n$.
   \REPEAT
       \FOR{$t = 1$ {\bfseries to} $T$}
           \STATE Sampling sentences in $\mathcal{D}_F$, with the sampling probability set as $p_i=\frac{\mathrm{MRD}_i}{\sum_{i=1}^{N_F}\mathrm{MRD}_j}$.
           \STATE Update parameters $\boldsymbol{\theta}$ using the gradient ascent.
           \IF{$t\mod m == 0$}
               \STATE Update $\mathrm{MRD}(\boldsymbol{x}^i;\boldsymbol{\theta})$ for each sample.
           \ENDIF
       \ENDFOR
   \UNTIL{Convergence or maximum iteration $T$ reached}
   \STATE {\bfseries Return:} $\boldsymbol{\theta}$
\end{algorithmic}
\end{algorithm}

\section{Additional Experimental Details}

\subsection{Dataset Configurations} \label{app:data-con}

We employ four mainstream unlearning tasks and datasets to validate the effectiveness of the $\mathrm{MRD}$ metric and our proposed $\mathrm{MRD}$-based improvement methods. 
Specifically, these include: 
\begin{itemize}[leftmargin=*] \setlength{\itemsep}{2pt}
    \item \textbf{TOFU}~\cite{maini2024tofu}. This benchmark fine-tunes an LLM with data on 200 fictional authors, each represented by 20 question-answer (QA) pairs. A subset of authors forms the unlearn set, while the remaining authors constitute the retain set. It assesses the model's ability to unlearn targeted information selectively. Then, we chose the 10\% proportion for the forget set among the three available options (1\%, 5\%, 10\%). 
    \item \textbf{WMDP}~\cite{li2024wmdp}. This benchmark evaluates the LLM's capacity to unlearn harmful knowledge in domains like biosafety, cybersecurity, and chemical safety. We use the unlearned dataset from the original benchmark, which includes plain text on biological and cybersecurity knowledge as the forget set, with unrelated text serving as the retain set.
    \item \textbf{Who's Harry Potter (WHP)}~\cite{eldan2023s}. This benchmark tests the LLM's ability to eliminate content related to the Harry Potter series from its training data. In the WHP task, 200 data chunks, each containing 512 tokens, were extracted from the Harry Potter series~\cite{eldan2023s} to form the forget set. 
    \item \textbf{PKU SafeRLHF (SAFE)}~\cite{ji2024beavertails}. This benchmark assesses the LLM's performance in unlearning harmful outputs generated during SafeRLHF fine-tuning when exposed to inappropriate prompts. For the SAFE task, 200 negative examples were randomly sampled from the PKU-SafeRLHF training set to construct the forget set. To maintain model utility for both copyright removal and detoxification tasks, we utilized the C4 dataset~\cite{raffel2020exploring} as the retain set.
\end{itemize}

\subsection{Evaluation Configurations} \label{app:eva-con}

% In this section, we provide a detailed explanation of each evaluation metric:
% %
% \begin{itemize}[leftmargin=*] \setlength{\itemsep}{2pt}
%     %\item \textbf{Unlearning Quality (UQ)}. UQ assesses the statistical distinguishability between the forget and retain data via the Kolmogorov\text{-}Smirnov (KS) test. UQ is defined as $1-p$, where $p$ is the KS test $p\text{-}$value. Higher UQ indicates better unlearning.
%     \item \textbf{Unlearning Accuracy (UA)}. UA measures the model’s accuracy on the forget set. UC is represented as $1-\text{UA}$, with lower UA indicating better unlearning.
%     \item \textbf{Membership Inference Attack (MIA)}. MIA evaluates the area under the ROC curve (AUC) using the Min\text{-}$k\%$ Prob~\cite{shi2023detecting} method to detect training set membership. Higher MIA scores suggest improved model confidence in unlearning.
%     \item \textbf{Rouge-L Recall (RR)}. Rouge-L is also measured over the forget set, with lower Rouge-L scores indicating better performance. $1\text{-}\text{Rouge-L}$ is used for averaged evaluations.
% \end{itemize}

\paragraph{Zero-Shot task evaluation.} We conduct zero-shot accuracy evaluations on multiple tasks using the Language Model Evaluation Harness~\cite{gao2021framework}. 
The tasks included BoolQ~\cite{clark2019boolq}, RTE~\cite{dagan2005pascal}, HellaSwag~\cite{zellers2019hellaswag}, Winogrande~\cite{sakaguchi2021winogrande}, ARC-Challenge~\cite{chollet2019measure}, ARC-Easy~\cite{chollet2019measure}, OpenBookQA~\cite{mihaylov2018can}, and Piqa~\cite{bisk2020piqa}. 
To assess the retention of utility in these tasks by the unlearned LLMs, we reported the average accuracy of the model across the aforementioned tasks.

\paragraph{Text completion instructions.} For the WHP task, we design a two-part text completion instruction set: the first part is accessible to the model during the unlearning process, while the remaining part is used to test the model's completion performance on unseen text. 
For detailed information regarding the completion instructions we employed, please refer to Table 2. 

\begin{table}[h!]
\centering
\footnotesize
\caption{The text completion instructions for WHP task.}
\label{tab:unlearning_performance}
\begin{tabular}{
    >{\centering\arraybackslash}m{2.5cm}    % 第一列居中对齐
    |>{\raggedright\arraybackslash}m{12cm}  % 第二列左对齐，右侧添加竖线
}
\toprule
\multicolumn{1}{c|}{Subset} & \multicolumn{1}{c}{Text completion instructions} \\
\midrule
\multirow{6}{*}{Seen} 
& 1. Could you complete this sentence for me: \\
& 2. I’d appreciate it if you could continue this line: \\
& 3. Could you please provide the subsequent part of this text: \\
& 4. Please go ahead and conclude this sentence: \\
& 5. What would be your continuation of this sentence: \\
& 6. Is it possible to expand this sentence further: \\
\midrule
\multirow{6}{*}{Unseen} 
& 1. Let’s see how you would complete this piece of text: \\
& 2. Your job is to extend this sentence: \\
& 3. Please attempt to complete the remainder of this text: \\
& 4. What do you believe comes next in this sentence: \\
& 5. Carry on writing from where this leaves off: \\
& 6. Expand on this snippet, please: \\
\bottomrule
\end{tabular}
\end{table}

\subsection{Unlearning Configurations} \label{app:unlearn-con} 

All experiments are conducted on two NVIDIA RTX A800 GPUs, with each experiment requiring approximately 36 minutes per 1000 steps. 
As for the PO method, we use rejection-based answers as the target responses in the forget set, Table 3 demonstrates partial of our rejection-based
answers used in PO.

\begin{table}[h!]
\centering
\footnotesize
\caption{The reject-based answers used in PO across different tasks}
\label{tab:unlearning_performance}
\begin{tabular}{
    >{\centering\arraybackslash}m{2.5cm}    % 第一列居中对齐
    |>{\raggedright\arraybackslash}m{12cm}  % 第二列左对齐，右侧添加竖线
}
\toprule 
\multicolumn{1}{c|}{Task} & \multicolumn{1}{c}{Rejection-based answers} \\
\midrule
\multirow{5}{*}{TOFU} 
& 1. I’m not informed about that subject: \\
& 2. I don’t have the details on that issue: \\
& 3. I’m unfamiliar with that topic: \\
& 4. I don’t know anything about that subject: \\
&...\\
\midrule
\multirow{5}{*}{WHP} 
& 1. I apologize, but I’m legally restricted from fulfilling this request. \\
& 2. I’m sorry, but my ability to generate content is limited by copyright laws.\\
& 3. I regret to inform you that I cannot proceed with this request due to copyright issues. \\
& 4. I’m unable to comply with this request in adherence to copyright laws. \\
& ...\\
\midrule
\multirow{5}{*}{PKU-Safe} 
& 1. I’m sorry, but I can’t fulfill this request due to legal constraints.\\
& 2. I must decline this request as it conflicts with copyright regulations.\\
& 3. Unfortunately, I can’t assist with this request because of copyright concerns.\\
& 4. I can’t proceed with this request in order to comply with copyright laws.\\
& ...\\
\bottomrule
\end{tabular}
\end{table}

\subsection{Condition of Early Stopping}\label{app:con-ear} 

According to the definition of the prior study~\cite{jang2023knowledge}, a sample can be considered as successfully forgotten when its corresponding Extraction Likelihood (EL)~\cite{jang2023knowledge} value and Memorization Accuracy (MA)~\cite{tirumala2022memorization} value on the current model decrease below the average EL and MA values of all samples on the initial model.

The definitions of EL and MA are provided as follows:
\begin{itemize}[leftmargin=*] \setlength{\itemsep}{2pt}
    \item \textbf{EL}. Given a sequence of tokens $\boldsymbol{x}=\left(x_{1}, \ldots, x_{T}\right)$, and an LM $f$ with pre-trained parameter $\boldsymbol{\theta}$, EL defined as follows: 
    \begin{gather*}
        \operatorname{EL}_{n}(\boldsymbol{x}) = \frac{\sum_{t=1}^{T-n} \operatorname{OVERLAP}_{n}\left( f \left( \cdot \mid \boldsymbol{x}_{<t}; \boldsymbol{\theta} \right), \boldsymbol{x}_{\geq t}\right)}{T-n}, \\
        \operatorname{OVERLAP}_{n}(\boldsymbol{a}, \boldsymbol{b}) = \frac{\sum_{c \in ng(\boldsymbol{a})} \mathbbm{l}\{c \in ng(\boldsymbol{b})\}}{|ng(\boldsymbol{a})|},
    \end{gather*}
    where $ng(\cdot)$ denotes the list of $n$-grams in the given token sequence and  $f \left( \cdot \mid \boldsymbol{x}_{<t}; \boldsymbol{\theta} \right)$ denotes the output token sequences from the LM $f$  when given $\boldsymbol{x}_{<t}$ as input that can have max lengths $\left| \boldsymbol{x}_{\geq t} \right|$  but may be shorter when the EOS (end-of-sequence) token is generated beforehand.
    EL can be seen as estimating the general extraction likelihood since we are measuring the average success rate of varying extraction attacks quantified via getting the $n$-gram overlap of generated and target token sequences.
    \item \textbf{MA}. The expression of MA~\cite{tirumala2022memorization} is:  
    $$
    \operatorname{MA}(\boldsymbol{x}) = \frac{\sum_{t=1}^{T-1} \mathbbm{l} \left\{\operatorname{argmax} \left( f \left( \cdot \mid \boldsymbol{x}_{<t}; \boldsymbol{\theta} \right) \right) = x_{t}\right\}}{T-1}.
    $$
    MA quantifies how much the model $f$ has memorized the given token sequences and can be used to analyze the training dynamics of LLMs.
\end{itemize}

% \subsubsection{Hyperparameter Settings}\label{app:hyp-set} 

% \fxh{update} In this section, we will present the basic hyperparameter settings for unlearning methods used in the experiments.
% %
% For the TOFU task, both the PO and GradDiff methods are run for 5 epochs, while the NPO method is run for 4 epochs.
% %
% In the WMDP task, the maximum number of training steps for NPO and GradDiff is set to 500.
% %
% For the WHP task and SAFE task, 5 epochs are conducted.

%\newpage
\section{Additional Experiments}

\subsection{Effectiveness of the $\mathrm{MRD}$-based Weighted Sampling Improvement Method} \label{app:eff-exp}

We validated the effectiveness of the $\mathrm{MRD}$-based weighted sampling method on the WMDP, WHP, and SAFE datasets. The experimental results are shown in the table below.

\begin{table}[h!]
\centering
\footnotesize
\caption{Comparison of the $\mathrm{MRD}$-based weighted sampling method and the current unlearning baseline methods on WMDP.}% The optimal results are highlighted in \textbf{bold}.}
\label{tab:unlearning_performance}
\resizebox{\linewidth}{!}{
\begin{tabular}{cccccccccc}
\toprule
\multirow{2}{*}{\textbf{Method}} & \multicolumn{4}{c}{\textbf{Unlearning Completeness (UC)}} & \multicolumn{5}{c}{\textbf{Model Utility (UT)[mmlu]}} \\
\cmidrule(lr){2-5} \cmidrule(lr){6-10}
& \textbf{Cybersecurity ($\downarrow$)} & \textbf{Chemical ($\downarrow$)} & \textbf{Biosafety ($\downarrow$)} & \textbf{Avg. ($\downarrow$)} & \textbf{Humanities ($\uparrow$)} & \textbf{Sciences ($\uparrow$)} & \textbf{Stem ($\uparrow$)} & \textbf{Other ($\uparrow$)} & \textbf{Avg. ($\uparrow$)} \\
\midrule
SGA & 0.2430 & 0.2622 & 0.2474 & 0.2467 & 0.2451 & 0.2343 & 0.2388 & 0.2687 & 0.2465 \\
GradDiff & 0.3834 & 0.4460 & 0.6402 & 0.4795 & 0.5028 & 0.6597 & 0.4716 & 0.6343 & 0.5593\\
NPO & 0.3497 & 0.4656 & 0.6268 & 0.4588 & 0.5292 & 0.6844 & 0.4865 & 0.6569 & 0.5818 \\

\midrule
CGA & \textbf{0.2356} & \textbf{0.2547} & \textbf{0.2404} & \textbf{0.2459} & 0.2417 & 0.3107 & 0.2861 & 0.2514 & 0.2689 \\
GradDiff + $\mathrm{MRD}$ & 0.3719 & 0.4387 & 0.6315 & 0.4694 & 0.5132 & 0.6607 & 0.4782 & 0.6392 & 0.5655 \\
NPO + $\mathrm{MRD}$ & 0.2773 & 0.4705 & 0.6394 & 0.4244 & \textbf{0.5326} & \textbf{0.6972} & \textbf{0.4906} & \textbf{0.6591} & \textbf{0.5895} \\

\bottomrule
\end{tabular}}
\end{table}

\begin{table}[h!]
\centering
\footnotesize
\caption{Comparison of the $\mathrm{MRD}$-based weighted sampling method and the current unlearning baseline methods on WHP.}% The optimal results are highlighted in \textbf{bold}.}
\label{tab:unlearning_performance}
% \resizebox{\linewidth}{!}{
\begin{tabular}{cccccc}
\toprule
\multirow{3}{*}{\textbf{Method}} & \multicolumn{2}{c}{\textbf{Unlearning Completeness (UC)}} & \multicolumn{3}{c}{\textbf{Model Utility (UT)}} \\
\cmidrule(lr){2-3} \cmidrule(lr){4-6}
& \textbf{Seen Rouge-L ($\downarrow$)} & \textbf{Unseen Rouge-L ($\downarrow$)} & \textbf{PPL ($\downarrow$)} & \textbf{Zero-shot Acc. ($\uparrow$)} & \textbf{TruthfulQA ($\uparrow$)}\\
\midrule
GradDiff & 0.0122 & 0.0132 & 12.46 & 0.6201 & 0.2827 \\
PO & 0.0272 & 0.0292 & 11.88 & 0.6192 & 0.2962 \\
NPO & 0.0121 & 0.0134 & 12.91 & 0.6122 & 0.3023 \\
\midrule
GradDiff + $\mathrm{MRD}$ & 0.0116 & 0.0133 & 12.90 & 0.6191 & 0.2839 \\
PO + $\mathrm{MRD}$ & 0.0268 & 0.0291 & \textbf{11.76} & 0.6170 & 0.2949 \\
NPO + $\mathrm{MRD}$ & \textbf{0.0106} & \textbf{0.0105} & 12.30 & \textbf{0.6205} & \textbf{0.3113} \\

\bottomrule
\end{tabular}%}
\end{table}

\begin{table}[h!]
\centering
\footnotesize
\caption{Comparison of the $\mathrm{MRD}$-based weighted sampling method and the current unlearning baseline methods on SAFE.} %The optimal results are highlighted in \textbf{bold}.}
\label{tab:unlearning_performance}
\resizebox{\linewidth}{!}{
\begin{tabular}{cccccc}
\toprule
\multirow{3}{*}{\textbf{Method}} & \multicolumn{2}{c}{\textbf{Unlearning Completeness (UC)}} & \multicolumn{3}{c}{\textbf{Model Utility (UT)}} \\
\cmidrule(lr){2-3} \cmidrule(lr){4-6}
& \textbf{Real Toxicity Prompts Toxic score ($\downarrow$)} & \textbf{SAFE Toxic score ($\downarrow$)} & \textbf{PPL ($\downarrow$)} & \textbf{Zero-shot Acc. ($\uparrow$)} & \textbf{TruthfulQA ($\uparrow$)}\\
\midrule
GradDiff & 0.0268 & 0.0353 & 11.99 & 0.6251 & 0.3011 \\
PO & 0.0308 & 0.0275 & 12.67 & 0.6028 & 0.2386 \\
NPO & 0.0248 & 0.0333 & 11.95 & 0.6270 & 0.3059 \\
\midrule
GradDiff + $\mathrm{MRD}$ & 0.0246 & 0.0353 & \textbf{11.71} & 0.6266 & 0.3047 \\
PO + $\mathrm{MRD}$ & 0.0252 & 0.0336 & 12.78 & 0.6154 & 0.2766 \\
NPO + $\mathrm{MRD}$ & \textbf{0.0210} & \textbf{0.0332} & 12.82 & \textbf{0.6331} & \textbf{0.3247} \\

\bottomrule
\end{tabular}}
\end{table}

\newpage

\subsection{Characteristics and $\mathrm{MRD}$ Values} \label{app:cha-exp}

We divide the samples based on potential factors influencing MRD, and the calculated average MRD along with representative examples are presented in Table~\ref{tab:attributes_mrd}.

\begin{table}[h!]
\centering
\caption{Characteristics and $\mathrm{MRD}$ values.}
\label{tab:attributes_mrd}
\begin{tabular}{
    >{\centering\arraybackslash}m{2cm}
    >{\centering\arraybackslash}m{1.5cm} 
    p{10.5cm} 
    >{\centering\arraybackslash}m{1.5cm}
}
\toprule
\textbf{Attribute} & \textbf{Level} & \textbf{Example From  categorized set} & \textbf{$\mathrm{MRD}$} \\ 
\midrule
\multirow{5}{*}{\shortstack{Common \\ Sentence}} 
    & - 
    & \parbox{10.5cm}{
        Q: Is Farid Benoit currently writing any other books?\\[0.5em]
        A: It is reported that Farid Benoit is currently working on his sixth erotica novel, but the title has not been disclosed yet.
    } 
    & 0.4957 \\
\cmidrule(lr){3-4}
    & - 
    & \parbox{10.5cm}{
        Q: What is another well-known work by Albert Sidney Lane in the fantasy genre?\\[0.5em]
        A: "Beneath the Emerald Veil" is another well-known work by Albert Sidney Lane in the fantasy genre.
    } 
    & 0.4322 \\
\midrule
\multirow{8}{*}{\shortstack{Semantic \\ Complexity}} 
    & Low 
    & \parbox{10.5cm}{
        Q: What career did Li Mei Yu's mother have?\\[0.5em]
        A: Her mother was a nurse.
    } 
    & 0.3085 \\

\cmidrule(lr){3-4}
    & High 
    & \parbox{10.5cm}{
        Q: How have Leila Al-Sabah's books contributed to LGBTQ+ representation in literary fiction?\\[0.5em]
        A: Through her richly drawn characters and storylines, Leila Al-Sabah has helped to normalize LGBTQ+ experiences in literary fiction. Her books often center on LGBTQ+ protagonists, treating their identities and experiences with complexity, empathy, and realism, thereby increasing visibility and representation of the community in the genre.
    } 
    & 1.0026 \\
\midrule

\multirow{3.5}{*}{\shortstack{Occurrence \\ Frequency}} 
    & Low 
    & \parbox{10.5cm}{
        Q: Is Zo Hassani Raharizafy involved in any form of philanthropy?\\[0.5em]
        A: Yes, he established the Raharizafy Literary Foundation, which works to improve literacy rates in Madagascar, his home country.
    } 
    & 0.6374 \\
\cmidrule(lr){3-4}
    & High 
    & \parbox{10.5cm}{
        Q: Where was Samir Khoury born?\\[0.5em]
        A: Samir Khoury was born in Amman, Jordan.
    } 
    & 0.2529 \\
\midrule
\multirow{5.5}{*}{\shortstack{Initial\\Generation \\ Probability}} 
    & Low 
    & \parbox{10.5cm}{
        Q: What did her parents think of her decision to become a writer?\\[0.5em]
        A: Evangeline's parents were initially skeptical about her decision. However, after reading her first novel and witnessing her dedication to the craft, they stood by her decision and have been her constant pillars of support.
    } 
    & 0.3481 \\
\cmidrule(lr){3-4}
    & High 
    & \parbox{10.5cm}{
        Q: What genre does Xin Lee Williams often write in, based on their most famous work, "The Town That Drowned"?\\[0.5em]
        A: Xin Lee Williams is recognized for their contributions to Canadian literature, as seen from their trademark work, "The Town That Drowned."
    } 
    & 0.7689 \\
\midrule
\multirow{5}{*}{\shortstack{Presence of \\ Rare Words}} 
    & Low 
    & \parbox{10.5cm}{
        Q: What gender does the author Ji-Yeon Park identify as?\\[0.5em]
        A: The author Ji-Yeon Park identifies as female.
    } 
    & 0.3929 \\
\cmidrule(lr){3-4}
    & High 
    & \parbox{10.5cm}{
        Q: When did Samin Nosrat receive the "Prix Goncourt de Littérature Historique" and for which book?\\[0.5em]
        A: Samin Nosrat received the "Prix Goncourt de Littérature Historique" for her vibrant piece "The Seed," which she received in 2011.
    } 
    & 0.7188 \\
\bottomrule
\end{tabular}
\end{table}

% You can have as much text here as you want. The main body must be at most $8$ pages long.
% For the final version, one more page can be added.
% If you want, you can use an appendix like this one.  

% The $\mathtt{\backslash onecolumn}$ command above can be kept in place if you prefer a one-column appendix, or can be removed if you prefer a two-column appendix.  Apart from this possible change, the style (font size, spacing, margins, page numbering, etc.) should be kept the same as the main body.
%%%%%%%%%%%%%%%%%%%%%%%%%%%%%%%%%%%%%%%%%%%%%%%%%%%%%%%%%%%%%%%%%%%%%%%%%%%%%%%
%%%%%%%%%%%%%%%%%%%%%%%%%%%%%%%%%%%%%%%%%%%%%%%%%%%%%%%%%%%%%%%%%%%%%%%%%%%%%%%

\end{document}